\documentclass[11pt, letterpaper]{article}
\usepackage[utf8]{inputenc} %
\usepackage[T1]{fontenc}    %
\usepackage[margin=1in]{geometry}

\usepackage{hyperref}
\usepackage{url}
\usepackage{cite}

\usepackage{setspace}

\usepackage{natbib}

\usepackage{amsmath,amssymb,amsthm,amscd,amsfonts,dsfont,bm}
\usepackage{graphicx,fancyhdr,color}
\usepackage{caption}
\usepackage{subcaption}

\newlength\tindent
\usepackage[parfill]{parskip}

\usepackage{amsmath,amsfonts,bm,dsfont,amsthm}

\def\eqref#1{equation~\ref{#1}}

\def\1{\bm{1}}

\def\vc{{\bm{c}}}

\def\ve{{\bm{e}}}
\def\vf{{\bm{f}}}

\def\vh{{\bm{h}}}

\def\vp{{\bm{p}}}

\def\vv{{\bm{v}}}

\def\mA{{\bm{A}}}

\def\mC{{\bm{C}}}

\def\mH{{\bm{H}}}
\def\mI{{\bm{I}}}

\def\mT{{\bm{T}}}
\def\mU{{\bm{U}}}
\def\mV{{\bm{V}}}

\def\mLambda{{\bm{\Lambda}}}

\DeclareMathAlphabet{\mathsfit}{\encodingdefault}{\sfdefault}{m}{sl}
\SetMathAlphabet{\mathsfit}{bold}{\encodingdefault}{\sfdefault}{bx}{n}

\newcommand{\PP}{\mathbb P}

\newcommand{\BR}{\mathds 1}

\usepackage{tcolorbox}
\definecolor{grey}{rgb}{0.33, 0.33, 0.33}

\newcommand{\squishlist}{
\begin{list}{{{\small{$\bullet$}}}}
{\setlength{\itemsep}{1pt}      \setlength{\parsep}{0pt}
\setlength{\topsep}{-2pt}       \setlength{\partopsep}{0pt}
\setlength{\leftmargin}{1em} \setlength{\labelwidth}{1em}
\setlength{\labelsep}{0.5em} } }
\newcommand{\squishend}{  \end{list}  }

\makeatletter
\renewcommand*\env@matrix[1][*\c@MaxMatrixCols c]{%
  \hskip -\arraycolsep
  \let\@ifnextchar\new@ifnextchar
  \array{#1}}
\makeatother

\newcommand{\tar}{{^\circ}}
\newcommand{\DP}{\textsf{DP}}
\newcommand{\EO}{\textsf{EOd}}
\newcommand{\EOd}{\textsf{EOd}}
\newcommand{\EOp}{\textsf{EOp}}
\newcommand{\uni}{u}
\newcommand{\eraw}{\textsf{Err}^{\textsf{raw}}}
\newcommand{\ecal}{\textsf{Err}^{\textsf{cal}}}
\newcommand{\globalT}{\textsf{Global}}
\newcommand{\localT}{\textsf{Local}}
\newcommand{\base}{\textsf{Base}}
\newcommand{\soft}{\textsf{Soft}}

\usepackage{hyperref}
\usepackage{url}
\usepackage{booktabs}

\usepackage{graphicx,epsfig}
\usepackage{xcolor,enumerate}
\usepackage{multirow}
\usepackage{tcolorbox}
\usepackage{enumitem}
\usepackage{algorithm}
\usepackage{algorithmic}

\newcommand\independent{\protect\mathpalette{\protect\independenT}{\perp}}
\def\independenT#1#2{\mathrel{\rlap{$#1#2$}\mkern2mu{#1#2}}}

\newcommand{\para}[1]{{\vspace{2pt} \noindent \textbf{#1}
    \hspace{6pt}}}

\newcommand{\eg}{{\em e.g.\ }}
\newcommand{\ie}{{\em i.e.\ }}
\newcommand{\etc}{{\em etc.}}

\definecolor{mygray}{gray}{0.6}
\newcommand{\algcom}[1]{\textsl{\color{mygray}{\footnotesize #1}}}

\ifodd 1

\newcommand{\yl}[1]{\textbf{\color{red}(Yang: #1)}}
\newcommand{\zzw}[2]{\textbf{\color{blue}(Zhaowei: #1)}{\color{blue}~#2}}
\newcommand{\jk}[1]{\textbf{\color{green}(Jiankai: #1)}}
\newcommand{\kevin}[1]{\textcolor{cyan}{[Kevin: #1]}}
\newcommand{\clar}[1]{\textbf{\color{green}(NEED CLARIFICATION: #1)}}

\else

\newcommand{\com}[1]{}
\newcommand{\clar}[1]{}
\newcommand{\response}[1]{}
\newcommand{\yl}[1]{}
\newcommand{\zzw}[2]{}
\newcommand{\jk}[1]{}
\newcommand{\kevin}[1]{}
\fi

\definecolor{frenchblue}{rgb}{0.0, 0.45, 0.73}
\definecolor{princetonorange}{rgb}{1.0, 0.56, 0.0}
\definecolor{darkgreen}{rgb}{0.33, 0.42, 0.18}
\definecolor{darkpastelgreen}{rgb}{0.01, 0.75, 0.24}

\newcommand{\dgreen}[1]{\textbf{\color{darkpastelgreen}#1}}
\definecolor{mygray}{gray}{0.6}

\usepackage{tcolorbox}
\newtcolorbox{mybox}{colback=darkgreen!5!white,colframe=darkgreen!10!white,left=1mm,top=1mm,right=1mm,boxsep=0mm,bottom=1mm,width=\linewidth,before=\par,after=\par,
}

\theoremstyle{plain}
\newtheorem{theorem}{Theorem}[section]

\newtheorem{requirement}[theorem]{Requirement}
\newtheorem{lemma}[theorem]{Lemma}
\newtheorem{corollary}[theorem]{Corollary}
\theoremstyle{definition}
\newtheorem{definition}[theorem]{Definition}
\newtheorem{assumption}[theorem]{Assumption}
\theoremstyle{remark}

\definecolor{myorange}{RGB}{241,167,108}
\definecolor{myblue}{RGB}{94,167,243}
\definecolor{myred}{RGB}{255,69,0}
\definecolor{mygreen}{RGB}{50,205,50}
\definecolor{grey}{rgb}{0.33, 0.33, 0.33}

\title{Weak Proxies are Sufficient and Preferable for Fairness \\ with Missing Sensitive Attributes}

\author{Zhaowei Zhu\thanks{Equal contributions. Part of the work was done while Z. Zhu interned at ByteDance AI Lab.}~$^\diamond$,  Yuanshun Yao$^{*\mathsection}$, Jiankai Sun$^\mathsection$, Hang Li$^\mathsection$, and Yang Liu$^{\mathsection}$\thanks{Corresponding authors: Y. Liu and Z. Zhu.}\\
$^\diamond$University of California, Santa Cruz, \texttt{zwzhu@ucsc.edu} \\ $^\mathsection$ByteDance, \texttt{\{kevin.yao,jiankai.sun,lihang.lh,yangliu.01\}@bytedance.com}
}
\date{}

\begin{document}

\maketitle

\begin{abstract}

Evaluating fairness can be challenging in practice because the sensitive attributes of data are often inaccessible due to privacy constraints. 
The go-to approach that the industry frequently adopts is using \textit{off-the-shelf proxy models} to predict the missing sensitive attributes, \eg Meta~\citep{meta} and Twitter~\citep{belli2022county}. 
Despite its popularity, there are three important questions unanswered: (1) Is directly using proxies efficacious in measuring fairness? (2) If not, is it possible to accurately evaluate fairness using proxies only?  (3) Given the ethical controversy over inferring user private information, is it possible to only use weak (\ie inaccurate) proxies in order to protect privacy?
Our theoretical analyses show that directly using proxy models can give a false sense of (un)fairness. \textit{Second}, we develop an algorithm that is able to measure fairness (provably) accurately with only three properly identified proxies.
\textit{Third}, we show that our algorithm allows the use of only weak proxies (\eg with only $68.85\%$ accuracy on COMPAS), adding an extra layer of protection on user privacy. Experiments validate our theoretical analyses and show our algorithm can effectively measure and mitigate bias. Our results imply a set of practical guidelines for practitioners on how to use proxies properly. Code is available at \url{github.com/UCSC-REAL/fair-eval}.
\end{abstract}

\section{Introduction}
The ability to correctly measure a model's fairness is crucial to studying and improving it %
\citep{corbett2018measure,madaio2022assessing,barocas2021designing}. %
However in practice it can be challenging since measuring group fairness requires access to the sensitive attributes of the samples,
which are often unavailable due to privacy regulations~\citep{andrus2021we,holstein2019improving,veale2017fairer}. For instance, the most popular type of sensitive information is demographic information. In many cases, it is unknown and illegal to collect or solicit. The ongoing trend of privacy regulations will further worsen the challenge. %

One straightforward solution is to use \textit{off-the-shelf proxy or proxy models} to predict the missing sensitive attributes. For example, Meta~\citep{meta} measures racial fairness by building proxy models to predict race from zip code based on US census data. Twitter employs a similar approach~\citep{belli2022county}.
This solution has a long tradition in other areas, \eg health~\citep{elliott2009using}, finance~\citep{baines2014fair}, and politics~\citep{imai2016improving}. It has become a standard practice and widely adopted in the industry due to its simplicity. %

\begin{figure}[!tb]
  \centering
  \includegraphics[width=0.5\linewidth]{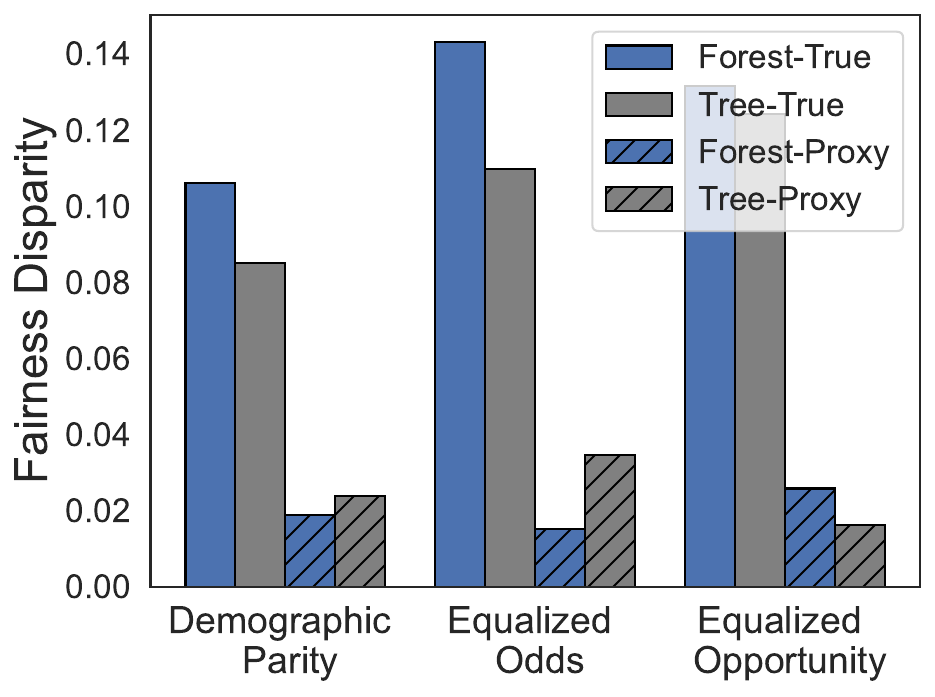}
  \caption{
  Fairness disparities of models on COMPAS \citep{angwin2016machine}.
  \emph{True (or Proxy)}: Disparities using ground-truth sensitive attribute values (or proxy model's predictions). \emph{Forest (or Tree)}: Random forest (or decisions tree) models. Observations: 1) Models considered as fair according to proxies can be actually unfair (True vs. Proxy), giving a false sense of fairness.
2) Fairness misperception (Forest vs. Tree) can cause practitioners to deploy wrong models.
  }
  \label{fig:motivation}

\end{figure}

Despite the popularity of this simple approach, few prior works have studied the efficacy or considered the practical constraints imposed by ethical concerns. \textit{In terms of efficacy}, it remains unclear to what degrees we can trust a reported fairness measure based on proxies. A misleading fairness measure can trigger decline of trust and legal concerns. Unfortunately, this indeed happens frequently in practice. For example, Figure~\ref{fig:motivation} shows the estimated fairness vs. true fairness on COMPAS~\citep{angwin2016machine} dataset with race as the sensitive attribute. We use proxy models to predict race from last name. 
There are two observations: 1) Models considered as fair according to proxies are actually unfair. The \emph{Proxy} fairness disparities ($0.02$) can be much smaller than the \emph{True} fairness disparities ($>0.10$), giving a false sense of fairness. 
2) Fairness misperception can be misleading in model selection. The proxy disparities mistakenly indicate random forest models have smaller disparities (\DP\ and \EO) than decision tree models, but in fact it is the opposite.

\textit{In terms of ethical concerns}, there is a growing worry on using proxies to infer sensitive information without user consent~\citep{twitterres,fosch2021gendering,leslie2019understanding,kilbertus2017avoiding}. Not unreasonably argued, using highly accurate proxies would reveal user's private information. We argue that practitioners should use inaccurate or \textit{weak proxies} whose noisy predictions would add additional protection to user privacy. 
However, if we merely compute fairness in the traditional way, the inaccuracy would propagate from weak proxies to the measured fairness metrics. To this end, we desire an algorithm that uses weak proxies only but can still accurately measure fairness.

We ask three questions: 
(1) Is directly using proxies efficacious in measuring fairness? 
(2) If not, is it possible to accurately evaluate fairness using proxies only?  (3) Given the ethical controversy over inferring user private information, is it possible to only use weak proxies to protect privacy?

We address those questions as follows:
\squishlist
\item\emph{Directly using proxies can be misleading:} We theoretically show that directly using proxy models to estimate fairness would lead to a fairness metric whose estimation can be off by a quantity proportional to the prediction error of proxy models and the true fairness disparity (Theorem~\ref{thm:baseline_error}, Corollary~\ref{coro:exact_base_error}).

\begin{figure*}[!t]
  \centering
  \includegraphics[width=0.95\linewidth]{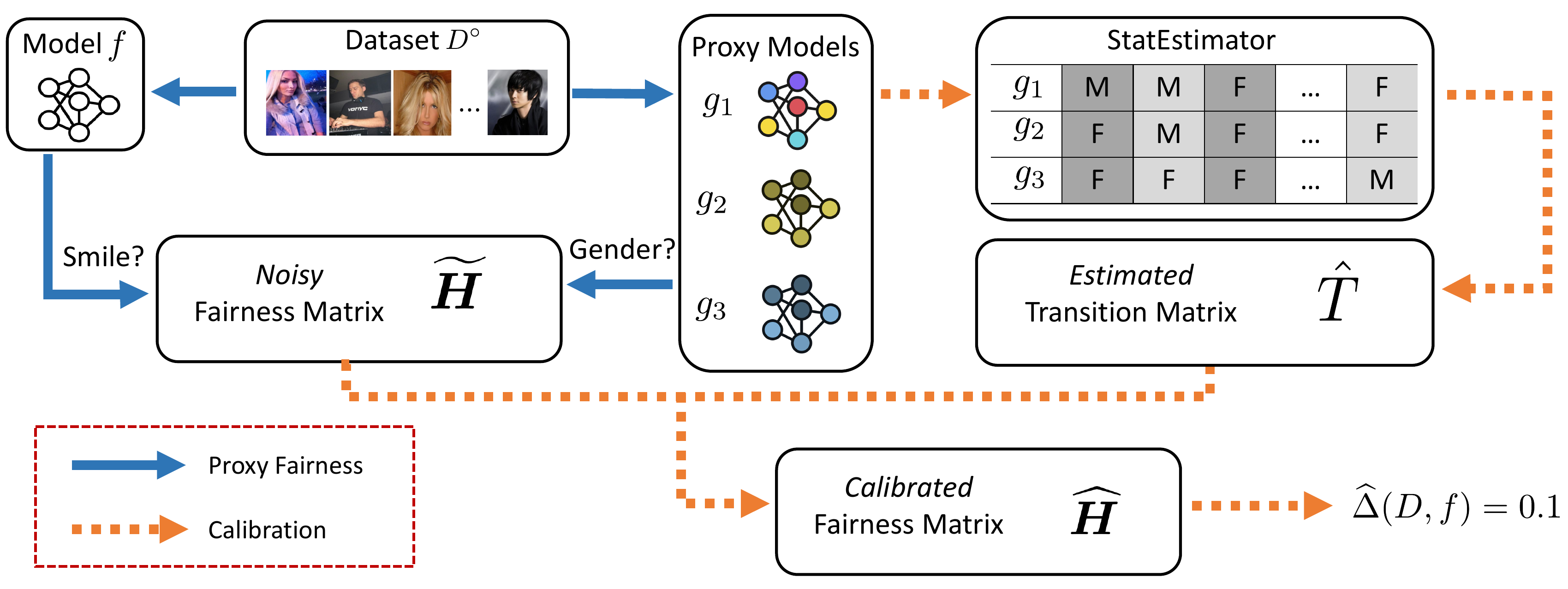}
  \caption{Overview of our algorithm that estimates fairness using only weak proxy models. We first directly estimate the noisy fairness matrix with proxy models (\textcolor{frenchblue}{blue arrows}), and then calibrate the estimated fairness matrix (\textcolor{princetonorange}{orange arrows}).
  }
  \label{fig:overview}
\end{figure*}

\item\emph{Provable algorithm using only weak proxies:} We propose an algorithm (Figure~\ref{fig:overview}, Algorithm~\ref{alg:alg_calibrate}) to calibrate the fairness metrics. We prove the error upper bound of our algorithm (Theorem~\ref{thm:error_correction}, Corollary~\ref{coro:compare}).
We further show \textit{three weak} proxy models with certain desired properties are sufficient and necessary to give unbiased fairness estimations using our algorithm (Theorem~\ref{thm:optimality}).

\item \emph{Practical guidelines:} We provide a set of practical guidelines to practitioners, including when to directly use the proxy models, when to use our algorithm to calibrate, how many proxy models are needed, and how to choose proxy models.

\item \emph{Empirical studies:} Experiments on COMPAS and CelebA consolidate our theoretical findings and show our calibrated fairness is significantly more accurately than baselines. We also show our algorithm can lead to better mitigation results.
\squishend

The paper is organized as follows. Section \ref{sec:pre} introduces necessary preliminaries. Section~\ref{sec:base_error} analyzes what happens when we directly use proxies, and shows it can give misleading results, which motivates our algorithm. Section~\ref{sec:method} introduces our algorithm that only uses weak proxies and instructions on how to use it optimally. Section~\ref{sec:exp} shows our experimental results. Section~\ref{sec:related} discusses related works and Section~\ref{sec:conclusion} concludes the paper.

\section{Preliminaries}\label{sec:pre}

Consider a $K$-class classification problem and a dataset %
$D{\tar} := \{(x_n, y_n)|n\in[N]\}$, where $N$ is the number of instances, $x_n$ is the \emph{feature}, and $y_n$ is the \emph{label}. Denote by $\mathcal X$ the feature space, $\mathcal Y=[K]:=\{1,2,\cdots, K\}$ the label space, and $(X,Y)$ the random variables of $(x_n,y_n),\forall n$.
The target model $f: \mathcal X \rightarrow [K]$ maps $X$ to a predicted label class $f(X)\in [K]$. We aim at measuring group fairness conditioned on a sensitive attribute $A\in [M] :=\{1,2,\cdots, M\}$ which is unavailable in $D{\tar}$. Denote the dataset with ground-truth sensitive attributes by $D:=\{(x_n, y_n,a_n)|n\in[N]\}$, the joint distribution of $(X,Y,A)$ by $\mathcal D$. 
The task is to estimate the fairness metrics of $f$ on $D\tar$ \emph{without} sensitive attributes such that the resulting metrics are as close to the fairness metrics evaluated on $D$ (with true $A$) as possible. 
We provide  a summary of notations in Appendix~\ref{app:notations}. 

We consider three group fairness definitions and their corresponding measurable metrics: \textit{demographic parity} (\DP) \citep{calders2009building,chouldechova2017fair}, \textit{equalized odds} (\EO) \citep{woodworth2017learning}, and \textit{equalized opportunity} (\EOp) \citep{hardt2016equality}.
All our discussions in the main paper are specific to \DP\ defined as follows but we include the \textit{complete derivations} for \EO\ and \EOp\ in Appendix.

\begin{definition}[Demographic Parity]\label{defn:dp} 
The demographic parity metric of $f$ on $\mathcal D$ conditioned on $A$ is defined as:
\vspace{-5pt}
\begin{align*}
 \Delta^{\DP}(\mathcal D, f) := \frac{1}{M(M-1)K} \cdot
\sum_{\substack{a,a' \in [M] \\ k\in[K]}}| \PP(f(X)=k|A=a) - \PP(f(X)=k|A=a')|.
\end{align*}
\end{definition}

\para{Matrix-form Metrics.}
For later derivations, we define matrix $\mH$ as an intermediate variable. Each column of $\mH$ denotes the probability needed for evaluating fairness with respect to $f(X)$.
For \DP, $\mH$ is a $M\times K$ matrix with $
H{[a,k]}:=\PP(f(X) = k| A=a).
$
The $a$-th row, $k$-th column, and $(a,k)$-th element of $\mH$ are denoted by $\mH[a],\mH[:,k]$, and $\mH[a,k]$, respectively.
Then $\Delta^{\DP}(\mathcal D, f)$ in Definition~\ref{defn:dp} can be rewritten as:
\begin{definition}[\DP\ - Matrix Form]\label{defn:dp-mat} 
\begin{align*}
\vspace{-5pt}
& \Delta^{\DP}(\mathcal D, f) := \frac{1}{M(M-1)K} \sum_{\substack{a,a' \in [M] \\ k\in[K]}}| \mH[a,k] - \mH[a',k]|.
\end{align*}
\vspace{-20pt}
\end{definition}
See definitions for \EO\ and \EOp\ in Appendix~\ref{appendix:more_fairness_def}.

\para{Proxy Models.}
The conventional way to measure fairness is to approximate $A$ with an proxy model $g: \mathcal X \rightarrow [M]$ \citep{ghazimatin2022measuring,awasthi2021evaluating,chen2019fairness} and get proxy (noisy) sensitive attribute $\widetilde A:=g(X)$\footnote{The input of $g$ can be any subsets of feature $X$. We write the input of $g$ as $X$ just for notation simplicity.}.

\para{Transition Matrix.}
The relationship between $\mH$ and $\widetilde\mH$ is largely dependent on the relationship between $A$ and $\widetilde A$ because it is the only variable that differs. 
Define matrix $\mT$ to be the transition probability from $A$ to $\widetilde A$ where $(a,\tilde a)$-th element is $T[a,\tilde a] = \PP(\widetilde A = \tilde a | A = a)$. 
Similarly, denote by $\mT_k$ the \emph{local} transition matrix conditioned on $f(X)=k$, where the $(a,\tilde a)$-th element is $T_{k}[{a,\tilde a}] := \PP( \widetilde A = \tilde a | f(X)=k , A = a)$.
We further define clean (\ie ground-truth) prior probability of $A$ as $\vp:=[\PP(A=1),\cdots, \PP(A=M)]^\top$ and the noisy (predicted by proxies) prior probability of $\widetilde A$ as $\tilde \vp:=[\PP(\widetilde A=1), \cdots, \PP(\widetilde A=M)]^\top$.

\section{Proxy Results Can be Misleading}
\label{sec:base_error}

This section provides an analysis on how much the measured fairness-if using proxies naively-can deviate from the reality.

\para{Using Proxy Models Directly.}
Consider a scenario with $C$ proxy models denoted by the set $\mathcal G:=\{g_1,\cdots, g_C\}$. The noisy sensitive attributes are denoted as $\widetilde A_c:=g_c(X),\forall c\in[C]$ and the corresponding target dataset with $\widetilde A$ is $\widetilde D:=\{(x_n,y_n, (\tilde a_n^1,\cdots,\tilde a_n^C) ) | n\in[N]\}$, drawn from a distribution $\widetilde{\mathcal D}$.
Similarly, by replacing $A$ with $\widetilde A$ in $\mH$, we can compute $\widetilde\mH$, which is the matrix-form noisy fairness metric estimated by the proxy model $g$ (or $\mathcal G$ if multiple proxy models are used). %
Define the directly measured fairness metric of $f$ on $\widetilde{\mathcal D}$ as follows.
\begin{definition}[Proxy Disparity - \DP]\label{defn:noisy_fair}
\begin{align*}
& \Delta^{\DP}(\widetilde{\mathcal D}, f) := \frac{1}{M(M-1)K} \sum_{\substack{a,a' \in [M] \\ k\in[K]}}| \widetilde\mH[a,k] -\widetilde \mH[a',k]|.
\end{align*}
\end{definition}

\para{Estimation Error Analysis.}
We study the error of proxy disparity and give practical guidelines implied by analysis.

Intuitively, the estimation error of proxy disparity depends on the error of the proxy model $g$.
Recall $\vp$, $\tilde\vp$, $\mT$ and $\mT_k$ are clean prior, noisy prior, global transition matrix, and local transition matrix. Denote by $\bm\mLambda_{\tilde \vp}$ and $\bm\mLambda_{\vp}$ the square diagonal matrices constructed from $\tilde \vp$ and $\vp$. We formally prove the upper bound of estimation error for the directly measured metrics in Theorem~\ref{thm:baseline_error} (See Appendix~\ref{app:theo1} for the proof).

\begin{theorem}[Error Upper Bound of Proxy Disparities]\label{thm:baseline_error}
Denote by $\eraw:= |\widetilde\Delta^{\DP}(\widetilde{\mathcal D}, f) - \Delta^{\DP}(\mathcal D, f)|$ the estimation error of the proxy disparity. Its upper bound is:
\begin{mybox}
{\begin{align*}
\eraw
\le  \frac{2}{K} \sum_{k\in[K]}  \Big(\bar h_{k}  \underbrace{\|\mLambda_{\tilde\vp} (\mT^{-1}\mT_k - \mI)\mLambda_{\tilde \vp}^{-1} \|_{1}}_{\text{cond. indep. violation}}  + \delta_k \underbrace{\|\mLambda_{\vp} \mT_{k} \mLambda_{\tilde\vp}^{-1}  - \mI\|_{1}}_{\text{error of $g$}} \Big),
\end{align*}
}
\end{mybox}
where
$\bar h_{k}:= \frac{1}{M}\sum\limits_{a\in[M]} H[a,k]$, $\delta_k := \max\limits_{a\in[M]}~|H[a,k] - \bar h_k|$.
\end{theorem}
It shows the error of proxy disparity depends on:
\squishlist
\item $\bar h_k$: The average confidence of $f(X)$ on class $k$ over all sensitive groups. For example, if $f$ is a crime prediction model and $A$ is race, a biased $f$ \citep{angwin2016machine} may predict that the crime ($k=1$) rate for different races are 0.1, 0.2 and 0.6 respectively, 
then $\bar h_1 = \frac{0.1+0.2+0.6}{3} = 0.3$, and it is an approximation (unweighted by sample size) of the average crime rate over the entire population. The term depends on $\mathcal D$ and $f$ only (\ie the true fairness disparity), and independent of any estimation algorithm. %
\item $\delta_k$: The maximum disparity between confidence of $f(X)$ on class $k$ and average confidence $\bar h_k$ across all sensitive groups. Using the same example, $\delta_1 = \max(|0.1 - 0.3|, |0.2 - 0.3|, |0.6 - 0.3|) = 0.3$.
It is an approximation of the underlying fairness disparity, and larger $\delta_k$ indicates $f$ is more biased on $\mathcal D$. The term is also dependent on $\mathcal D$ and $f$ (\ie the true fairness disparity), and independent of any estimation algorithm.
\item %
\emph{Conditional Independence Violation}: The term is dependent on the proxy model $g$'s prediction $\tilde A$ in terms of the transition matrix ($\mT$ and $\mT_k$) and noisy prior probability ($\tilde \vp$). The term goes to $0$ when $\mT = \mT_k$, which implies $\tilde A$ and $f(X)$ are independent conditioned on $A$. This is the common assumption made in the prior work~\citep{awasthi2021evaluating,prost2021measuring,fogliato2020fairness}. And this term measures how much the conditional independence assumption is violated. 

\item \emph{Error of $g$}: %
The term depends on the proxy model $g$. It goes to $0$ when $\mT_k = \mI$ which implies the error rates of $g$'s prediction is $0$, \ie $g$ is perfectly accurate. It measures the impact of $g$'s error on the fairness estimation error.
\squishend

\para{Case Study.} To help better understand the upper bound, we consider a simplified case when both $f$ and $A$ are binary. We further assume the conditional independence condition to remove the third term listed above in Theorem~\ref{thm:baseline_error}. See Appendix~\ref{app:cond_indep} for the formal definition of conditional independence.\footnote{We only assume it for the purpose of demonstrating a less complicated theoretical result, we do \emph{not} need this assumption in our proposed algorithm later.}
Corollary~\ref{coro:exact_base_error} summarizes the result.

\begin{corollary}\label{coro:exact_base_error}
For a binary classifier $f$ and a binary sensitive attribute $A \in \{1, 2\}$, when ($\tilde{A} \independent f(X) | A$) holds, Theorem~\ref{thm:baseline_error} is simplified to 
\begin{mybox}
$$\eraw \le  {\delta}  \left( \PP( A=1 | \widetilde A=2) + \PP( A=2 | \widetilde A=1)\right),$$
\end{mybox}
where 
$\delta = |\PP(f(X)=1|A=1) - \PP(f(X)=1|A=2)|$.
\end{corollary}

Corollary~\ref{coro:exact_base_error} shows the estimation error of proxy disparity is proportional to the true underlying disparity between sensitive groups (\ie $\delta$) and the proxy model's error rates. In other words, the uncalibrated metrics can be highly inaccurate when $f$ is highly biased or $g$ has poor performance. This leads to the following suggestions:

\para{Guidelines for Practitioners.} We should only trust the estimated fairness from proxy models when (1) the proxy model $g$ has good performance \textit{and} (2) the true disparity is small\footnote{Of course, without true sensitive attribute we can never know for sure how large the true disparity is. But we can infer it roughly based on problem domain and known history. For example, racial disparity in hiring is known to exist for a long time. We only need to know if the disparity is extremely large or not.} (\ie the target model $f$ is not highly biased).

In practice, both conditions required to trust the proxy results are frequently violated. When we want to measure $f$'s fairness, often we already have some fairness concerns and therefore the underlying fairness disparity is unlikely to be negligible. And the proxy model $g$ is usually inaccurate due to privacy concerns (discussed in Section~\ref{sec:req}) and distribution shift. This motivates us to develop an approach for more accurate estimates.

\section{Weak Proxies Suffice}\label{sec:method}
In this section, we show that by properly using a set of proxy models, we are able to guarantee an unbiased estimate of the true fairness measures.

\subsection{Proposed Algorithm}\label{sec:framework}
With a given proxy model $g$ that labels sensitive attributes, we can anatomize the relationship between the true disparity and the proxy disparity.
The following theorem targets \DP\  and see Appendix~\ref{appendix:lemma_relation_true_noisy} for results with respect to \EO\ and \EOp\ and their proofs.
\begin{theorem}\label{lem:relation_true_noisy}
[Closed-form Relationship (\DP)]
The closed-form relationship between the true fairness vector $\mH[:,k]$ and the noisy fairness vector $\widetilde \mH[:,k]$ is the following:
\begin{mybox}
\[
\mH[:,k] = (\mT_k^\top\mLambda_{\vp})^{-1}\mLambda_{\tilde{\vp}}\widetilde \mH[:,k], \forall k\in[K].
\]
\end{mybox}

\end{theorem}

\para{Insights.} Theorem~\ref{lem:relation_true_noisy} reveals that the proxy disparity and the corresponding true disparity are related in terms of  three key statistics: noisy prior $\tilde\vp$, clean prior $\vp$, and local transition matrix $\mT_k$.
Ideally, if we have the ground-truth values of them, we can calibrate the noisy fairness vectors to their corresponding ground-truth vectors (and therefore obtaining the perfectly accurate fairness metrics) usingTheorem~\ref{lem:relation_true_noisy}.
Hence, the most important step is to estimate $\mT_k$, $\vp$, and $\tilde\vp$ without knowing the ground-truth values of $A$. Once we have those estimated key statistics, we can easily plug them into the above equation as the calibration step. Figure~\ref{fig:overview} shows the overview of our algorithm.

\begin{algorithm}[!t]
\caption{Fairness calibration algorithm (\DP)
}\label{alg:alg_calibrate} 
\begin{algorithmic}[1]
\STATE \textbf{Input:} A set of proxy models $\mathcal G = \{g_1, \cdots, g_C \}$. Target dataset $D\tar$. Target model $f$. 
Transition matrix and prior probability estimator \texttt{StatEstimator}.
\\
\algcom{\# Predict sensitive attributes using all $g\in\mathcal G$}
\STATE $\tilde a_n^c \leftarrow g_c(x_n), \forall c\in[C], n\in[N]$
\\
\algcom{\# Build the dataset with noisy sensitive attributes}
\STATE $\widetilde D \leftarrow \{(x_n, y_n, (\tilde a_n^1,\cdots, \tilde a_n^C)) | n\in[N]\}$ 
\\
\algcom{ \# Estimate fairness matrix and prior with sample mean}
\STATE $\widetilde\mH, \tilde\vp \leftarrow \texttt{DirectEst}(\widetilde D,f)$ %
\\
\algcom{\# Estimate key statistics: $\vp$ and $\mT_k$}
\STATE $\{\widehat\mT_1, \cdots, \widehat\mT_K\}, \hat\vp \leftarrow \texttt{StatEstimator}(\widetilde D, f)$ %
\\
\algcom{ \# Calibrate each fairness vector with Theorem~\ref{lem:relation_true_noisy}}
\STATE  $\forall k\in[K]: \widehat\mH[:,k] \leftarrow (\widehat\mT_k^\top\mLambda_{\hat\vp})^{-1}\mLambda_{\tilde{\vp}}\widetilde \mH[:,k]$
\\
\algcom{ \# Calculate the final fairness metric as Definition~\ref{defn:dp-mat}}
\STATE {\small $\widehat{\Delta}(\widetilde{D}, f) \leftarrow \frac{1}{M(M-1)K} \sum_{\substack{a,a' \in [M] \\ k\in[K]}}| \widehat\mH[a,k] - \widehat\mH[a',k]|.
$}
\STATE 
\textbf{Output:} The calibrated fairness metric $\widehat{\Delta}(\widetilde{D}, f)$

\end{algorithmic}
\end{algorithm}

\para{Algorithm.} We summarize the method in Algorithm~\ref{alg:alg_calibrate}. In Line~4, we use sample mean in the uncalibrated form to estimate $\widetilde\mH$ as $\widetilde H[\tilde a,k] = \PP(f(X)=k|\widetilde A=\tilde a) \approx \frac{1}{N} \sum_{n=1}^N\BR(f(x_n=k|\tilde a_n=\tilde a))$ and $\tilde\vp$ as $ \tilde \vp[\tilde a] = \PP(\widetilde A=\tilde a) \approx \frac{1}{N}\sum_{n=1}^N\BR(\tilde a_n=\tilde a)$, $\forall \tilde a\in[M]$.
In Line~5, we plug in an existing transition matrix and prior probability estimator to estimate $\mT_k$ and $\vp$ with only mild adaption that will be introduced shortly.
Note that although we choose a specific estimator, our algorithm is a flexible framework that is compatible with any \texttt{StatEstimator} proposed in the noisy label literature~\citep{liu2017machine,zhu2021clusterability,zhu2022beyond}.

\para{Details: Estimating Key Statistics.}
The algorithm requires us to estimate $\mT_k$ and $\vp$ based on the predicted $\widetilde A$ by proxy models. In the literature of noisy learning, there exists several feasible algorithms \citep{liu2015classification,scott2015rate,patrini2017making,northcutt2021confident,zhu2021clusterability}. We choose HOC \citep{zhu2021clusterability} because it has stronger theoretical guarantee and lower sample complexity than most existing estimators. Intuitively, if given three proxy models, the joint distributions of their predictions would encode $\mT_k$ and $\vp$, \ie $\PP(\widetilde A_1, \widetilde A_2, \widetilde A_3) = \textsf{Func}(\{\mT_k\}_{k\in[K]}, \vp)$.
For example, with Chain rule and independence among proxy predictions conditioned on $A$,  we have:
\begin{align*}
    \PP(\widetilde A_1 = \tilde a_1, \widetilde A_2 = \tilde a_2, \widetilde A_3 =\tilde a_3 | f(X) = k)  
    = \sum_{a \in [M]} \PP(A=a | f(X)=k) \cdot T_k[a, \tilde a_1] \cdot T_k[a, \tilde a_2] \cdot T_k[a, \tilde a_3].
\end{align*}

HOC counts the frequency of different $(\widetilde A_1, \widetilde A_2, \widetilde A_3)$ patterns to obtain LHS and solve equations to get $T_k$'s in the RHS. 
See more details of our implementations in Appendix~\ref{appendix:hoc_statestimator}--\ref{appendix:hoc_general} and HOC in Appendix~\ref{sec:hoc_detail}.

\subsection{Requirements of Proxy Models}

\label{sec:req}
To use our algorithm, there are two practical questions for practitioners: 1) what properties proxy models should satisfy and 2) how many proxy models are needed. The first question is answered by two requirements made in the estimation algorithm HOC:

\begin{requirement}[Informativeness of Proxies]\label{ap:informative}
The noisy attributes given by each proxy model $g$ are informative, \ie  $\forall k\in[M]$, 1) $\mT_k$ is non-singular and 2) either $T_k[a,a] > \PP(\widetilde A=a|f(X)=k)$
or $T_k[a,a] > T_k[a,a'], \forall a'\ne a$.
\end{requirement}
Requirement~\ref{ap:informative} is the prerequisite of getting a feasible and unique estimate of $\mT_k$  \citep{zhu2021clusterability}, where the non-singular requirement ensures the matrix inverse in Theorem~\ref{lem:relation_true_noisy} exists and the constraints on $T_k[a,a]$ describes the worst tolerable performance of $g$. When $M=2$, the constraints can be simplified as $T_k[1,2] + T_k[2,1] < 1$ \citep{liu2017machine,liu2019peer}, \ie $g$'s predictions are better than random guess in binary classification. If this requirement is violated, there might exist more than one feasible estimates of $\mT_k$, making the problem insoluble. 

The above requirement is weak. The proxies are merely required to positively correlate with the true sensitive attributes. We discuss the privacy implication of using weak proxies shortly after.

\begin{requirement}[Independence between Proxies]\label{ap:iid}
The noisy attributes predicted by proxy models $g_1(X), \cdots, g_C(X)$  are independent and identically distributed (i.i.d.) given $A$. %
\end{requirement}
Requirement~\ref{ap:iid} ensures the additional two proxy models provide more information than using only one classifier. If it is violated, we would still get an estimate but may be inaccurate. Note this requirement is different from the conditional independence often assumed in the fairness literature~\citep{awasthi2021evaluating,prost2021measuring,fogliato2020fairness}, which is $g(X) \independent f(X) | A$ rather than ours $g_1(X) \independent g_2(X) \independent g_3(X) | A$.

The second question (how many proxy models are needed) has been answered by Theorem~5 in \citet{liu2022identifiability}, which we summarize in the following.
\begin{lemma}\label{lem:num_labelers}
If satisfying Requirements~\ref{ap:informative}--\ref{ap:iid}, three proxy models are both sufficient and necessary to identify $\mT_k$.  
\end{lemma}

\para{How to Protect Privacy with Weak Proxies.} Intuitively, weak proxies can protect privacy better than strong proxies due to their noisier predictions. We connect weak proxy's privacy-preserveness to \textit{differential privacy}~\citep{ghazi2021deep}. Assume misclassification probability on $\widetilde{A}$ is bounded across all samples, \ie $$\max_{x \in X}{\PP(\widetilde A = a | A = a, X = x)} \le 1 - \epsilon_0$$ and $$\min_{x \in X}{\PP(\widetilde A = a | A = a', X = x)} \ge  \epsilon_1$$ for $a\in[M], a'\in[M], a\ne a'$. Then the proxy predictions satisfy $\ln(\frac{1-\epsilon_0}{\epsilon_1})$-DP\footnote{We mean the definition of \textit{label differential privacy}~\citep{ghazi2021deep} that studies privacy of label of proxy models, which is the sensitive attribute $A$. We avoid using the term LabelDP or LDP to not confuse with label $Y$.}. See Appendix~\ref{appendix:labeldp} for the proof.

In practice, if the above assumption does not hold naturally by proxies, we can add noise to impose it. When practitioners think proxies are too strong, they can add additional noise to reduce informativeness, further protecting privacy. Later we will show in Table~\ref{table:celebA_noise_facenet512} that our algorithm is robust in estimation accuracy when adding noise to proxy predictions. When we intentionally make the proxies weaker by flipping predicted sensitive attributes with probability $0.4$, resulting in only $58.45\%$ proxy accuracy, it corresponds to $0.41$-DP ($\epsilon_0 = \epsilon_1 = 0.4$) protection.

\subsection{Theoretical Guarantee}\label{sec:thm_our_method}
We theoretically analyze estimation error on our calibrated metrics in a similar way as in Section~\ref{sec:base_error}. 
Denote by $\widehat\Delta^{\DP}(\widetilde{\mathcal D}, f)$ the calibrated DP disparity evaluated on our calibrated fairness matrix $\widehat\mH$. We have:
\begin{theorem}[Error Upper Bound of Calibrated Metrics]\label{thm:error_correction}
Denote the estimation error of the calibrated fairness metrics by $\ecal:= |\widehat\Delta^{\DP}(\widetilde{\mathcal D}, f) - \Delta^{\DP}(\mathcal D, f)|$. Then:
\begin{mybox}
\begin{align*}
\ecal
\le  \frac{2}{K}\sum_{k\in[K]} \left\|\mLambda_{{\vp}}^{-1} \right\|_1  \left\|\mLambda_{{\vp}} \mH[:,k]  \right\|_\infty \varepsilon(\widehat\mT_{k},\hat \vp),
\end{align*}
\end{mybox}
where $\varepsilon(\widehat\mT_k,\hat \vp) := \| \mLambda_{{\hat\vp}}^{-1}\mLambda_{{\vp}} - \mI\|_{1}\|\mT_{k}\widehat\mT_k^{-1}\|_{1} + \|  \mI -  \mT_{k}\widehat\mT_k^{-1}  \|_{1}$ is the error induced by calibration. With a perfect estimator $\widehat\mT_k=\mT_k$ and $\hat\vp_k=\vp_k, \forall k\in[K]$, we have $\ecal=0$. 
\end{theorem}

Theorem~\ref{thm:error_correction} shows the upper bound of estimation error mainly depends on the estimates $\widehat \mT_k$ and $\hat\vp$, \ie  the following two terms in  $\varepsilon(\widehat\mT_k,\hat \vp)$: $\| \mLambda_{{\hat\vp}}^{-1}\mLambda_{{\vp}} - \mI \|_{1}\|\mT_{k}\widehat\mT_k^{-1}\|_{1}$ and   $\|  \mI -  \mT_{k}\widehat\mT_k^{-1}  \|_{1}$. When the estimates are perfect, \ie $\widehat\mT_k = \mT_k$ and $\hat \vp = \vp$, then both terms go to 0 because $\mLambda_{{\hat\vp}}^{-1}\mLambda_{{\vp}} = \mI$ and $\mT_{k}\widehat\mT_k^{-1} = \mI$. 
Together with Lemma~\ref{lem:num_labelers}, we can show the optimality of our algorithm as follows.

\begin{theorem}
\label{thm:optimality}
When Requirements~\ref{ap:informative}--\ref{ap:iid} hold for three proxy models, the calibrated fairness metrics given by Algorithm~\ref{alg:alg_calibrate} with key statistics estimated by Algorithm~\ref{alg:alg_hoc} achieve zero error, \ie $|\widehat\Delta^{\DP}(\widetilde{\mathcal D}, f) - \Delta^{\DP}(\mathcal D, f)| = 0$.
\end{theorem}

Besides, we compare the error upper bound of our method with the exact error (not its upper bond) in the case of Corollary~\ref{coro:exact_base_error}, and summarize the result in Corollary~\ref{coro:compare}.

\begin{corollary}\label{coro:compare}
For a binary classifier $f$ and a binary sensitive attribute $A \in \{1, 2\}$, when ($\tilde{A} \independent f(X) | A$) and $\vp = [0.5,0.5]^\top$,
the proposed calibration method is guaranteed to be more accurate than the uncalibrated measurement, \ie, $\ecal \le \eraw$, if
\begin{mybox}
$$
    \varepsilon(\widehat\mT_{k},\hat \vp) \le \gamma:= \max\limits_{k'\in\{1,2\}} \frac{  e_1 + e_2}{ 1 + \frac{\|\mH[:,k']\|_1}{\Delta^{\DP}(\mathcal D,f)}}, \forall k \in \{1,2\}.
$$
\end{mybox}
\end{corollary}
Corollary~\ref{coro:compare} shows when the error $\varepsilon(\widehat\mT_{k},\hat \vp)$ that is induced by inaccurate $\widehat\mT_k$ and $\hat\vp$ is below the threshold $\gamma$, our method is guaranteed to lead to a smaller estimation error compared to the uncalibrated measurement under the considered setting. The threshold implies that, adopting our method rather than the uncalibrated measurement can be greatly beneficial when $e_1$ and $e_2$ are high (\ie $g$ is inaccurate) \textit{or} when the normalized (true) fairness disparity $\frac{\Delta^{\DP}(\mathcal D,f)}{\|\mH[:,k']\|_1}$ is high (\ie $f$ is highly biased).

\subsection{Guidelines for Practitioners}
\label{sec4:guide}
We provide a set of guidelines implied by our theoretical results.

\para{When to Use Our Algorithm.} Corollary~\ref{coro:compare} shows that our algorithm is preferred over directly using proxies when 1) the proxy model $g$ is weak \textit{or} 2) the true disparity is large.

\para{How to Best Use Our Algorithm.} Section~\ref{sec:req} implies a set of principles for selecting proxy models:
\squishlist
\item[i)] [Requirement~\ref{ap:informative}] Even if proxy models are weak, as long as they are informative, \eg in binary case the performance is better than random guess, then it is enough for estimations.
\item[ii)] [Requirement~\ref{ap:iid}] We should try to make sure the predictions of proxy models are i.i.d., which is more important than using more proxy models. One way of doing it is to choose proxy models trained on different data sources. 
\item[iii)] [Lemma~\ref{lem:num_labelers}] At least three proxy models are prefered.
\squishend

\begin{table*}[!t]
	\caption{Normalized estimation error on COMPAS. True disparity: $\sim 0.2$. Average accuracy of weak proxy models: \dgreen{68.85\%}.}
	\begin{center}
		\scalebox{0.87}{\footnotesize{\begin{tabular}{c|cc cc| cccc | cccc} 
			\toprule
			 COMPAS  &  \multicolumn{4}{c}{\emph{\DP\ Normalized Error (\%) $\downarrow$}} & \multicolumn{4}{c}{\emph{\EO\ Normalized Error (\%) $\downarrow$}} & \multicolumn{4}{c}{\emph{\EOp\ Normalized Error (\%) $\downarrow$}} \\
			 Target models $f$ & \base & \soft & \globalT & \localT & \base & \soft & \globalT & \localT& \base & \soft & \globalT & \localT\\
			  			\midrule \midrule
tree & 43.82 & 61.26 & \textbf{22.29} & 39.81 & 45.86 & 63.96 & \textbf{23.09} & 42.81 & 54.36 & 70.15 & \textbf{13.27} & 49.49 \\
forest & 43.68 & 60.30 & \textbf{19.65} & 44.14 & 45.60 & 62.85 & \textbf{18.56} & 44.04 & 53.83 & 69.39 & \textbf{17.51} & 63.62 \\
boosting & 43.82 & 61.26 & \textbf{22.29} & 44.64 & 45.86 & 63.96 & \textbf{23.25} & 49.08 & 54.36 & 70.15 & \textbf{13.11} & 54.67 \\
SVM & 50.61 & 66.50 & \textbf{30.95} & 42.00 & 53.72 & 69.69 & \textbf{32.46} & 47.39 & 59.70 & 71.12 & \textbf{29.29} & 51.31 \\
logit & 41.54 & 60.78 & \textbf{16.98} & 35.69 & 43.26 & 63.15 & \textbf{21.42} & 31.91 & 50.86 & 65.04 & \textbf{14.90} & 26.27 \\
nn & 41.69 & 60.55 & \textbf{19.48} & 34.22 & 43.34 & 62.99 & \textbf{19.30} & 43.24 & 54.50 & 68.50 & \textbf{14.20} & 59.95 \\
compas\_score & 41.28 & 58.34 & \textbf{11.24} & 14.66 & 42.43 & 59.79 & \textbf{11.80} & 18.65 & 48.78 & 62.24 & \textbf{5.78} & 23.80 \\
		   \bottomrule
		\end{tabular}}}
	\end{center}
	\label{table:compas}
\end{table*}

\begin{table*}[!t]
	\caption{Normalized error on CelebA. We simulate weak proxies by adding noise to predicted attributes according to Requirement~\ref{ap:iid} to bring down the performance of proxy models. Each row represents the noise magnitude and accuracy of proxy models, \eg ``$[0.2, 0.0]$ (82.44\%)'' means $T[1,2] = 0.2$, $T[2,1] = 0.0$ and accuracy is \dgreen{82.44\%}.
	}
	\begin{center}
		\scalebox{0.87}{\footnotesize{\begin{tabular}{c|cc cc| cccc | cccc} 
			\toprule
			 CelebA   &  \multicolumn{4}{c}{\emph{\DP\ Normalized Error (\%) $\downarrow$}} & \multicolumn{4}{c}{\emph{\EO\ Normalized Error (\%) $\downarrow$}} & \multicolumn{4}{c}{\emph{\EOp\ Normalized Error (\%) $\downarrow$}} \\
			  FaceNet512 & \base & \soft & \globalT & \localT & \base & \soft & \globalT & \localT& \base & \soft & \globalT & \localT\\
			  			\midrule \midrule
    $[0.2, 0.0]$ (\dgreen{82.44\%}) & 7.37 & 11.65 & 20.58 & \textbf{5.05} & 25.06 & 26.99 & 6.43 & \textbf{0.10} & 24.69 & 27.27 & 11.11 & \textbf{1.07} \\
    $[0.2, 0.2]$ (\dgreen{75.54\%}) & 30.21 & 31.57 & 24.25 & \textbf{13.10} & 44.73 & 46.36 & 11.26 & \textbf{9.04} & 37.67 & 38.77 & \textbf{20.94} & 27.98 \\
    $[0.4, 0.2]$ (\dgreen{65.36\%}) & 51.32 & 54.56 & 19.42 & \textbf{10.47} & 62.90 & 65.10 & \textbf{11.09} & 19.15 & 56.51 & 58.73 & 23.86 & \textbf{23.55} \\
    $[0.4, 0.4]$ (\dgreen{58.45\%}) & 77.76 & 78.39 & \textbf{9.41} & 19.80 & 79.31 & 80.10 & 24.49 & \textbf{8.02} & 78.35 & 79.62 & 10.61 & \textbf{5.71} \\
		   \bottomrule
		\end{tabular}}}
	\end{center}
	\label{table:celebA_noise_facenet512}
\end{table*}

\section{Experiments}\label{sec:exp}

\subsection{Setup}\label{sec:exp_setting}
We test the performance of our method on two real-world datasets: COMPAS \citep{angwin2016machine} and CelabA \citep{liu2015faceattributes}. We report results on all three group fairness metrics (\DP, \EO, and \EOp) whose true disparities (estimated using the ground-truth sensitive attributes) are denoted by $\Delta^{\DP}(\mathcal D,f)$, $\Delta^{\EO}(\mathcal D,f)$, $\Delta^{\EOp}(\mathcal D,f)$ respectively. We train the target model $f$ on the dataset without using $A$, and use the proxy models downloaded from open-source projects. The detailed settings are the following:

\squishlist
\item \textbf{COMPAS} \citep{angwin2016machine}: Recidivism prediction data. Feature $X$: tabular data. Label $Y$: recidivism within two years (binary). Sensitive attribute $A$: race (black and non-black).Target models $f$ (trained by us): decision tree, random forest, boosting, SVM, logit model, and neural network (accuracy range 66\%--70\% for all models). Three proxy models ($g_1,g_2,g_3$): racial classifiers given name as input \citep{sood2018predicting} (average accuracy 68.85\%).%
\item \textbf{CelabA} \citep{liu2015faceattributes}: Face dataset.  Feature $X$: facial images. Label $Y$: smile or not (binary). Sensitive attribute $A$: gender (male and female). Target models $f$: ResNet18 \citep{he2016deep} (accuracy 90.75\%, trained by us).
We use one proxy model ($g_1$): gender classifier that takes facial images as input~\citep{serengil2021lightface}, and then use the clusterability to simulate the other two proxy models (as Line~4 in Algorithm~\ref{alg:alg_hoc}, Appendix~\ref{appendix:hoc_statestimator}). Since the proxy model $g_1$ is highly accurate (accuracy 92.55\%), which does not give enough privacy protection, we add noise to $g_1$'s predicted sensitive attributes according to Requirement~\ref{ap:iid}. We generate the other two proxies ($g_2$ and $g_2$) based on $g_1$'s noisy predictions.

\squishend

\paragraph{Method.} We propose a simple heuristic in our algorithm to stabilize estimation error on $\widehat{\mT}_{k}$. Specifically, we use a single transition matrix $\widehat{\mT}$ estimated once on the full dataset $\widetilde D$ as Line~8 of Algorithm~\ref{alg:alg_hoc} (Appendix~\ref{appendix:hoc_statestimator}) to approximate ${\mT}_{k}$. We name this heuristic as \globalT\ (\ie $\mT_k\approx \widehat{\mT}$) and the original method (estimated on each data subset $\widetilde D_k:=\{(X, Y, A) | f(X) = k\}$, \ie $\mT_k \approx \widehat{\mT}_k$) as \localT. See Appendix~\ref{appendix:discuss_global} for details. 
We compare with two baselines: the directly estimated metric without any calibration (\base) and \soft\ \citep{chen2019fairness} which also only uses proxy models to calibrate the measured fairness by re-weighting metric with the soft predicted probability from the proxy model.

\para{Evaluation Metric.}
Let $\Delta(D,f)$ be the ground-truth fairness metric. For a given estimated metric $E$, we define three estimation errors: $\textsf{Raw Error}(E) := |E - \Delta(D,f)|,$ $\textsf{Normalized Error}(E):=\frac{\textsf{Raw Error(E)}}{{\Delta(D,f)}},$ and $\textsf{Improvement}(E) := 1 - \frac{\textsf{Raw Error(E)}}{\textsf{Raw Error(Base)}},$ where \base{} is the directly measured metric.

\vspace{-5pt}
\subsection{Results and Analyses}
\label{subsec:exp_results}
\vspace{-5pt}

\para{COMPAS Results.} Table~\ref{table:compas} reports the normalized error on COMPAS (See Table~\ref{table:compas_full} in Appendix~\ref{app:compas_full} for the other two evaluation metrics). There are two main observations. \textit{First}, our calibrated metrics outperform baselines with a big margin on all three fairness definitions. Compared to \base, our metrics are 39.6\%--88.2\% more accurate (\textsf{Improvement}). As pointed out by Corollary~\ref{coro:compare}, this is because the target models $f$ are highly biased (Table~\ref{table:disparity_compas}) and the proxy models $g$ are inaccurate (accuracy 68.9\%). As a result, \base{} has large normalized error (40--60\%). \textit{Second}, \globalT{} outperforms \localT, since with inaccurate proxy models, Requirements~\ref{ap:informative}--\ref{ap:iid} on HOC may not hold in local dataset, inducing large estimation errors in local estimates. Finally, we also include the results with three-class sensitive attributes (black, white, and others) in Appendix~\ref{appendix:compas_3}.

\para{CelebA Results.} Table~\ref{table:celebA_noise_facenet512} summarizes the key results (see Appendix~\ref{app:celeba_full} for the full results). \textit{First}, our algorithm outperform baselines significantly on all fairness definitions with all noise rates, which validates Corollary~\ref{coro:compare}. When $g$ becomes less accurate, \base's \DP\ normalized error increases by more than $10$x while our error (\localT) only increases by $3$x. \textit{Second}, unlike COMPAS, \localT{} now outperforms \globalT. This is because we add random noise following Requirement~\ref{ap:iid} and therefore the estimation error of \localT\ is not increased significantly. This further consolidates our theoretical findings. Therefore when Requirement~\ref{ap:iid} is satisfied, using \localT{} can give more accurate estimations than \globalT\  (see Appendix~\ref{appendix:discuss_global} for more discussions). In practice, practitioners can roughly examine Requirement~\ref{ap:iid} by running statistical tests like Chi-squared tests on proxy predictions.

\para{Mitigating Disparity.}
We further discuss the disparity mitigation built on our method. The aim is to improve the classification accuracy while ensuring fairness constraints. Particularly, we choose \DP\ and test on CelebA, where $\widehat \Delta^\DP(\widetilde D,f) = 0$ is the constraint for our method and $\widetilde \Delta^\DP(\widetilde D,f) = 0$ is the constraint for the baseline (\base). Recall $\widehat \Delta^\DP(\widetilde D,f)$ is obtained from Algorithm~\ref{alg:alg_calibrate} (Line~8), and $\widetilde D:=\{(x_n,y_n,\tilde a_n) | n\in[N]\}$. Table~\ref{table:mitigation} shows our methods with popular pre-trained feature extractors (rows other than \base) can consistently achieve both a lower \DP\ disparity and a higher accuracy on the test data.
Besides, our method can achieve the performance which is close to the mitigation with ground-truth sensitive attributes.
We defer more details to Appendix~\ref{appendix:mitigation}.

\begin{table}[!t]
	\caption{Results (averaged by the last 5 epochs) of disparity mitigation. \base: Direct mitigation using noisy sensitive attributes. Ground-Truth: Mitigation using ground-truth sensitive attributes. Facenet, Facenet 512, \etc: Pre-trained models to generate feature representations that we use to simulate the other two proxy models.
	}
	\vspace{5pt}
	\begin{center}
		\scalebox{0.89}{\footnotesize{\begin{tabular}{c|c c} 
			\toprule
			 CelebA 
			    & $\Delta^\DP(D^{\text{test}},f)$ $\downarrow$ & Accuracy $\uparrow$ \\
			  			\midrule \midrule
\base  & 0.0578 & 0.8422\\
Ground-Truth & 0.0213 & 0.8650 \\
Facenet  & 0.0453 & 0.8466\\
Facenet512  & 0.0273 & 0.8557\\
OpenFace  & 0.0153 & 0.8600\\
ArcFace  & 0.0435 & 0.8491 \\
Dlib &  0.0265 & 0.8522\\
SFace  & 0.0315 & 0.8568\\
		   \bottomrule
		\end{tabular}}}
	\end{center}
	\label{table:mitigation}
\end{table}

\para{Guidelines for Practitioners.} The above experimental results lead to the following suggestions:
\squishlist
\item[1)] Our algorithm can give a clear advantage over baselines when the proxy model $g$ is weak (\eg error $\ge 15\%$) or the target model $f$ is highly biased (\eg fairness disparity $\ge 0.1$).
\item[2)] When using our algorithm, we should prefer \localT{} when Requirement~\ref{ap:iid} is satisfied, \ie proxies make i.i.d predictions; and prefer \globalT{} otherwise. In practice, practitioners can use statistical tests like Chi-squared tests to roughly judge if proxy predictions are independent or not.
\squishend

\section{Related Work}
\label{sec:related}

\para{Fairness with Imperfect Sensitive Attributes.} Existing methods mostly fall into two categories. \textit{First}, some assume access to ground-truth sensitive attributes on a data subset or label them if unavailable, \eg Youtube asks its creators to voluntarily provide their demographic information~\citep{youtube}. But it either requires labeling resource or depends on the volunteering willingness, and it suffers from sampling bias. \textit{Second}, some works assume there exists proxy datasets that can be used to train proxy models, \eg Meta~\citep{meta} and others~\citep{elliott2009using,awasthi2021evaluating,diana2022multiaccurate}. However, they often assume proxy datasets and the target dataset are \textit{i.i.d.}, and some form of conditional independence, which can be violated in practice. In addition, since proxy datasets also contain sensitive information (\ie the sensitive labels), it might be difficult to obtain such training data from the open-source projects.
The closest work to ours is \citep{chen2019fairness}, which also assumes only proxy models. It is only applicable to \textit{demographic disparity}, and we compare it in the experiments. Note that compared to the prior works, our algorithm only requires realistic assumptions. Specifically we drop many commonly made assumptions in the literature, \ie 1) access to labeling resource \citep{youtube}, 2) access to proxy model's training data \citep{awasthi2021evaluating,diana2022multiaccurate}, 3) data \textit{i.i.d} \citep{awasthi2021evaluating}, and 4) conditional independence \citep{awasthi2021evaluating,prost2021measuring,fogliato2020fairness}.

\para{Noisy Label Learning.}
Label noise may come from various sources, e.g., human annotation error \citep{xiao2015learning,wei2022learning,agarwal2016learning} and model prediction error \citep{lee2013pseudo,berthelot2019mixmatch,zhu2021rich}, which can be characterized by transition matrix on label \citep{liu2022identifiability,bae2022noisy,yang2021estimating}. 
Applying the noise transition matrix to ensure fairness is emerging \citep{wang2021fair,liu2021can,lamy2019noise}.
There exist two lines of works for estimating transition matrix. The first line relies on anchor points (samples belonging to a class with high certainty) or their approximations \citep{liu2015classification,scott2015rate,patrini2017making,xia2019anchor,northcutt2021confident}. These works requires training a neural network on the $(X,\widetilde A:=g(X))$. The second line of work, which we leverage, is \emph{data-centric} \citep{liu2017machine,liu2020surrogate,zhu2021clusterability,zhu2022beyond} and training-free. The main idea is to check the agreements among multiple noisy attributes as discussed in Appendix~\ref{sec:hoc_detail}.

\section{Conclusions and Discussions}\label{sec:conclusion}
Although it is appealing to use proxies to estimate fairness when sensitive attributes are missing, its ethical implications are causing practitioners to be cautious about adopting this approach. However simply giving up this practical and powerful solution shuts down the chance of studying fairness on a large scale. In this paper, we have offered a middle-way solution, \ie by using only weak proxies, we can protect data privacy while still being able to measure fairness. To this end, we design an algorithm that, though only based on weak proxies, can still provably achieve accurate fairness estimations. We show our algorithm can effectively measure and mitigate bias, and provide a set of guidelines for practitioners on how to use proxies properly. We hope our work can inspire more discussions on this topic since the inability to access sensitive attributes ethically is currently a major obstacle to studying and promoting fairness.

\clearpage
\newpage
\bibliography{ref}
\bibliographystyle{abbrvnat}
\clearpage
\newpage
\appendix
\onecolumn
\section*{Ethics Statement}
Our goal is to better study and promote fairness. Without a promising estimation method, given the increasingly stringent privacy regulations, it would be difficult for academia and industry to measure, detect, and mitigate bias in many real-world scenarios. However, we need to caution readers that, needless to say, no estimation algorithm is perfect. Theoretically, in our algorithm, if the transition matrix is perfectly estimated, then our method can measure fairness with 100\% accuracy. However, if Requirements~\ref{ap:informative}--\ref{ap:iid} required by our estimator in Algorithm~\ref{alg:alg_hoc} do not hold, our calibrated metrics might have a non-negligible error, and therefore could be misleading. In addition, the example we use to explain terms in Theorem~\ref{thm:baseline_error} is based on conclusions from~\citep{angwin2016machine}. We do not have any biased opinion on the crime rate across different racial groups. Furthermore, we are fully aware that many sensitive attributes are not binary, \eg race and gender. We use the binary sensitive attributes in experiments because 1) existing works have shown that bias exists in COMPAS between race \emph{black} and others and 2) the ground-truth gender attribute in CelebA is binary. We also have experiments with three categories of races (\emph{black, white, others}) in Appendix~\ref{appendix:compas_3}. We summarize races other than \emph{black} and \emph{white} as \emph{others} since their sample size is too small. Finally, all the data and models we use are from open-source projects, and the bias measured on them do not reflect our opinions about those projects.

\vspace{20pt}

{\LARGE \bf Appendix}

The Appendix is organized as follows. 
\squishlist    
    \item Section~\ref{app:def_notations_ap} presents a summary of notations, more fairness definitions, and a clear statement of the assumption that is common in the literature. Note our algorithm does \emph{not} rely on this assumption. 
    \item Section~\ref{app:proof} presents the full version of our theorems (for \DP, \EO, \EOp), corollaries, and the corresponding proofs.
    \item Section~\ref{sec:discuss_T} shows how HOC works and analyzes why other learning-centric methods in the noisy label literature may not work in our setting.
    \item Section~\ref{app:exp} presents more experimental results.
\squishend

\section{More Definitions and Assumptions}\label{app:def_notations_ap}

\subsection{Summary of Notations}\label{app:notations}

\begin{table}[H]
    \centering
    \caption{Summary of key notations}
    \scalebox{0.67}{\begin{tabular}{c|l}
    \toprule
    Notation & Explanation \\
    \midrule
      $\mathcal G:=\{g_1,\cdots,g_C\}$ & $C$ proxy models for generating noisy sensitive attributes \\
      $X,Y,A,$ and $\widetilde A:=g(X)$ & Random variables of feature, label, ground-truth sensitive attribute, and noisy sensitive attributes \\
      $x_n,y_n,a_n$ & The $n$-th feature, label, and ground-truth sensitive attribute in a dataset \\
      $N, K, M$ & The number of instances, label classes, categories of sensitive attributes\\
      $[N]:=\{1,\cdots,N\}$ & A set counting from $1$ to $N$ \\
      $\mathcal X$, $f:\mathcal X\rightarrow [K]$ & Space of $X$, target model \\
      $D{\tar} := \{(x_n, y_n)|n\in[N]\}$ & Target dataset  \\
      $D := \{(x_n, y_n, a_n)|n\in[N]\}$ & $D\tar$ with ground-truth sensitive attributes  \\
      $\widetilde D := \{(x_n, y_n, (\tilde a_n^1,\cdots, \tilde a_n^C))|n\in[N]\}$ & $D\tar$ with noisy sensitive attributes  \\
      $(X,Y,A)\sim\mathcal D, (X,Y,\widetilde A)\sim\widetilde{\mathcal D}$ & Distribution of $D$ and $\widetilde D$ \\
      \midrule
      $\uni\in\{\DP, \EO, \EOp\}$ & A unified notation of fairness definitions, e.g., \EO, \EOp, \EOd \\
      $\Delta^\uni(\mathcal D,f), \widetilde\Delta^\uni(\widetilde{\mathcal D},f), \widehat\Delta^\uni(\widetilde{\mathcal D},f)$ & True, (direct) noisy, and calibrated group fairness metrics on data distributions \\
      $\Delta^\uni( D,f), \widetilde\Delta^\uni(\widetilde{ D},f), \widehat\Delta^\uni(\widetilde{ D},f)$ & True, (direct) noisy, and calibrated group fairness metrics on datasets\\
      $\mH, \mH[a], \mH[:,k], \mH[a,k]$  & Fairness matrix, its $a$-th row, $k$-th column, $(a,k)$-th element \\
      $\widetilde \mH$  & Noisy fairness matrix with respect to $\widetilde A$ \\
      $\mT$, $T[a,a\tilde]:=\PP(\widetilde A=\tilde a | A=a)$ & Global noise transition matrix \\ 
      $\mT_k$, $T_k[a,a\tilde]:=\PP(\widetilde A=\tilde a | A=a, f(X)=k)$ & Local noise transition matrix \\ 
      $\vp:=[\PP(A=1), \cdots, \PP(A=M)]^\top$ & Clean prior probability \\
      $\tilde\vp:=[\PP(\widetilde A=1), \cdots, \PP(\widetilde A=M)]^\top$ & Clean prior probability \\
    \bottomrule
    \end{tabular}}
    \label{tab:notations}
\end{table}

\subsection{More Fairness Definitions}\label{appendix:more_fairness_def}

We present the full version of fairness definitions and the corresponding matrix form for \DP, \EO, and \EOp\ as follows.

\para{Fairness Definitions.}
We consider three group fairness~\citep{wang2020robust,cotter2019optimization,chen2022fairness} definitions and their corresponding measurable metrics: \textit{demographic parity} (\DP) \citep{calders2009building,chouldechova2017fair}, \textit{equalized odds} (\EO) \citep{woodworth2017learning}, and \textit{equalized opportunity} (\EOp) \citep{hardt2016equality}.

\textbf{Definition 2.1} (Demographic Parity). 
The demographic parity metric of $f$ on $\mathcal D$ conditioned on $A$ is defined as:
\begin{align*}
& \Delta^{\DP}(\mathcal D, f) := \frac{1}{M(M-1)K} \cdot \\
&\sum_{\substack{a,a' \in [M] \\ k\in[K]}}| \PP(f(X)=k|A=a) - \PP(f(X)=k|A=a')|.
\end{align*}

\begin{definition}[Equalized Odds]\label{defn:eod} 
The equalized odds metric of $f$ on $\mathcal D$ conditioned on $A$ is:
\[
\Delta^{\EO}(\mathcal D, f) =  \frac{1}{M(M-1)K^2}   \sum_{\substack{a,a' \in [M] \\ k\in[K], y\in[K]}} | \PP(f(X)=k|Y=y,A=a) - \PP(f(X)=k|Y=y,A=a')|.
\]
\end{definition}
\begin{definition}[Equalized Opportunity]\label{defn:eop} 
The equalized opportunity metric of $f$ on $\mathcal D$ conditioned on $A$ is:
\[
\Delta^{\EOp}(\mathcal D, f) = \frac{1}{M(M-1)}   \sum_{a,a' \in [M]} | \PP(f(X)=1|Y=1,A=a) - \PP(f(X)=1|Y=1,A=a')|.
\]
\end{definition}
\para{Matrix-form Metrics.} To unify three fairness metrics in a general form, we represent them with a matrix $\mH$. Each column of $\mH$ denotes the probability needed for evaluating fairness with respect to classifier prediction $f(X)$.
For \DP, $\mH{[:,k]}$ denotes the following column vector:
\[\mH{[:,k]}:=[ \PP(f(X) = k| A=1),  \cdots, \PP(f(X) = k| A=M)]^\top.\]

Similarly for \EO\ and \EOp, let ${k \otimes y}:=K(k-1)+y$ be the 1-d flattened index that represents the 2-d coordinate in $f(X) \times Y$, $\mH{[:,{k \otimes y}]}$ is defined as the following column vector:
\[\mH{[:,{k \otimes y}]}:=[ \PP(f(X) = k|Y=y, A=1), \cdots, \PP(f(X) = k|Y=y, A=M)]^\top.\]
The sizes of $\mH$ for \DP{}, \EO\, and \EOp\ are $M \times K$, $M \times K^2$, and $M \times 1$ respectively.
The noise transition matrix related to \EO\ and \EOp\ is $\mT_{k\otimes y}$, where the $(a,\tilde a)$-th element is denoted by $T_{k\otimes y}[{a,\tilde a}] := \PP( \widetilde A = \tilde a | f(X)=k, Y=y, A = a)$.

\subsection{Common Conditional Independence Assumption in the Literature}
\label{app:cond_indep}

We present below a common conditional independence assumption in the literature~\citep{awasthi2021evaluating,prost2021measuring,fogliato2020fairness}. Note our algorithm successfully drops this assumption.
\begin{assumption}[Conditional Independence]\label{def:tl-ind}
$\tilde A$ and $f(X)$ are conditionally independent given $A$ (and $Y$ for \EO, \EOp):
\begin{align*}
  & \text{\DP: } \PP(\widetilde A=\tilde a | f(X) = k, A=a)  =  \PP(\widetilde A=\tilde a |A=a), \forall a,\tilde a\in[M], k\in[K].\\ 
  & (i.e.  \tilde{A} \independent f(X) | A). \notag \\
  & \text{\EO{} / \EOp: } \PP(\widetilde A=\tilde a | f(X) = k, Y=y, A=a)  =  \PP(\widetilde A=\tilde a |Y=y,A=a), \forall a,\tilde a\in[M], k,y\in[K]. \\
  & (i.e. \tilde{A} \independent f(X) | Y, A).\notag 
 \label{eq:dp:cond_ind}
\end{align*}
\end{assumption}

\newpage

\section{Proofs}\label{app:proof}

\subsection{Full Version of Theorem~\ref{thm:baseline_error} and Its Proof}
\label{app:theo1}

Denote by $\mT_y$ the attribute noise transition matrix with respect to label $y$, whose $(a,\tilde a)$-th element is $T_y[a,\tilde a]:=\PP(\widetilde A=\tilde a|A=a,Y=y)$. Note it is different from $\mT_k$.
Denote by $\mT_{{k \otimes y}}$ the attribute noise transition matrix when $f(X)=k$ and $Y=y$, where the $(a,\tilde a)$-th element is $\mT_{{k \otimes y}}[{a,\tilde a}] := \PP( \widetilde A = \tilde a | f(X)=k , Y=y, A = a)$. Denote by $\vp_y:=[\PP(A=1|Y=y), \cdots, \PP(A=K|Y=y) ]^\top$ and $\tilde{\vp}_y:=[\PP(\widetilde A=1|Y=y), \cdots, \PP(\widetilde A=K|Y=y) ]^\top$ the clean prior probabilities and noisy prior probability, respectively.

\textbf{Theorem 3.2} (Error Upper Bound of Noisy Metrics)
Denote by $\eraw_\uni:= |\Delta^{\uni}(\widetilde{\mathcal D}, f) - \Delta^{\uni}(\mathcal D, f)|$ the estimation error of the directly measured noisy fairness metrics. Its upper bound is:
\squishlist
\item \DP: 
\begin{align*}
\eraw_\DP
\le  \frac{2}{K}\sum_{k\in[K]} \left(\bar h_{k}  \underbrace{\|\mLambda_{\tilde\vp} (\mT^{-1}\mT_k - \mI)\mLambda_{\tilde \vp}^{-1} \|_{1}}_{\text{cond. indep. violation}} + \delta_k \underbrace{\|\mLambda_{\vp} \mT_{k} \mLambda_{\tilde\vp}^{-1}  - \mI\|_{1}}_{\text{error of $g$}} \right).
\end{align*}
where
$\bar h_{k}:= \frac{1}{M}\sum\limits_{a\in[M]} H[a,k]$, $\delta_k := \max\limits_{a\in[M]}~|H[a,k] - \bar h_k|$.

\item \EO:
\begin{align*}
\eraw_\EO
\le  \frac{2}{K^2}\sum_{k\in[K],y\in[K]} \left(\bar h_{k\otimes y}  \underbrace{\|\mLambda_{\tilde\vp_y} (\mT_y^{-1}\mT_{k\otimes y} - \mI)\mLambda_{\tilde \vp_y}^{-1} \|_{1}}_{\text{cond. indep. violation}} + \delta_{k\otimes y} \underbrace{\|\mLambda_{\vp_y} \mT_{k\otimes y} \mLambda_{\tilde\vp_y}^{-1}  - \mI\|_{1}}_{\text{error of $g$}} \right).
\end{align*}
where
$\bar h_{k \otimes y}:= \frac{1}{M}\sum\limits_{a\in[M]} H[a,k\otimes y]$, $\delta_{k\otimes y} := \max\limits_{a\in[M]}~|H[a,k\otimes y] - \bar h_{k\otimes y}|$.

\item \EOp:
We obtain the result for \EOp\ by simply letting $k=1$ and $y=1$, i.e.,
\begin{align*}
\eraw_\EOp
\le  2\sum_{k=1, y=1} \left(\bar h_{k\otimes y}  \underbrace{\|\mLambda_{\tilde\vp_y} (\mT_y^{-1}\mT_{k\otimes y} - \mI)\mLambda_{\tilde \vp_y}^{-1} \|_{1}}_{\text{cond. indep. violation}} + \delta_{k\otimes y} \underbrace{\|\mLambda_{\vp_y} \mT_{k\otimes y} \mLambda_{\tilde\vp_y}^{-1}  - \mI\|_{1}}_{\text{error of $g$}} \right).
\end{align*}
where
$\bar h_{k \otimes y}:= \frac{1}{M}\sum\limits_{a\in[M]} H[a,k\otimes y]$, $\delta_{k\otimes y} := \max\limits_{a\in[M]}~|H[a,k\otimes y] - \bar h_{k\otimes y}|$.
\squishend

\begin{proof}

The following proof builds on the relationship derived in the proof for Theorem~\ref{lem:relation_true_noisy}. We encourage readers to check Appendix~\ref{appendix:lemma_relation_true_noisy} before the following proof.

Recall $\mT_{y}[a,a'] := \PP(\widetilde A=a' | A=a, Y=y)$.
Note
\[
\mLambda_{\tilde \vp_{y}} \bm 1 = \mT_y^\top \mLambda_{\vp_{y}} \bm 1\Leftrightarrow
(\mT_y^\top)^{-1}\mLambda_{\tilde \vp_{y}} \bm 1 =  \mLambda_{\vp_{y}} \bm 1.
\]
Denote by
\[
\mH[:,{k \otimes y}] = \bar h_{{k \otimes y}} \bm 1 + \vv_{{k \otimes y}},
\]
where $\bar h_{{k \otimes y}}:= \frac{1}{M}\sum_{a\in[M]}\PP(f(X)=k|A=a,Y=y)$.
We have
\[
\mLambda_{\vp_{y}} \mH[:,{k \otimes y}] 
= \bar h_{{k \otimes y}} \mLambda_{\vp_{y}} \bm 1 + \mLambda_{\vp_{y}} \vv_{{k \otimes y}}
= \bar h_{{k \otimes y}} (\mT_y^\top)^{-1}\mLambda_{\tilde \vp_{y}} \bm 1 + \mLambda_{\vp_{y}} \vv_{{k \otimes y}}.
\]
We further have
\begin{align*}
    & \widetilde \mH[:{k \otimes y}] \\
=   & \left(\mLambda_{\tilde\vp_{y}}^{-1} \mT_{{k \otimes y}}^\top \mLambda_{\vp_{y}} - \mI\right) \mH[:,{k \otimes y}] +  \mH[:,{k \otimes y}] \\
=   & \bar h_{{k \otimes y}}  \mLambda_{\tilde\vp_{y}}^{-1} \mT_{{k \otimes y}}^\top  (\mT_y^\top)^{-1}\mLambda_{\tilde \vp_{y}} \bm 1 + \mLambda_{\tilde\vp_{y}}^{-1} \mT_{{k \otimes y}}^\top  \mLambda_{\vp_{y}} \vv_{{k \otimes y}} - \bar h_{{k \otimes y}} \bm 1 - \vv_{{k \otimes y}} +  \mH[:,{k \otimes y}] \\
=   & \bar h_{{k \otimes y}}  \mLambda_{\tilde\vp_{y}}^{-1} \left(\mT_{{k \otimes y}}^\top  (\mT_y^\top)^{-1} - \mI \right)\mLambda_{\tilde \vp_{y}} \bm 1 +  \left(\mLambda_{\tilde\vp_{y}}^{-1}\mT_{{k \otimes y}}^\top  \mLambda_{\vp_{y}} - \mI\right) \vv_{{k \otimes y}}  +  \mH[:,{k \otimes y}].
\end{align*}

Noting $|A| - |B| \le|A+B|\le |A| + |B|$, we have $|~|A+B| - |B|~|\le |A|$.
Therefore,
\begin{align*}
    &\left|~ \left|(\ve_{\tilde a} - \ve_{\tilde a'})^\top \widetilde \mH[:{k \otimes y}]\right| - \left|(\ve_{\tilde a} - \ve_{\tilde a'})^\top  \mH[:{k \otimes y}]\right|~ \right| \\
\le & \bar h_{{k \otimes y}}\left|(\ve_{\tilde a} - \ve_{\tilde a'})^\top   \mLambda_{\tilde\vp_{y}}^{-1} \left(\mT_y^{-1}\mT_{{k \otimes y}}   - \mI \right)^\top\mLambda_{\tilde \vp_{y}} \bm 1\right|  \qquad \text{(Term 1)} \\
    & + \left|(\ve_{\tilde a} - \ve_{\tilde a'})^\top    \left(\mLambda_{\tilde\vp_{y}}^{-1}\mT_{{k \otimes y}}^\top  \mLambda_{\vp_{y}} - \mI\right) \vv_{{k \otimes y}} \right|. \qquad \quad~~ \text{(Term 2)}\\
\end{align*}
Term-1 and Term-2 can be upper bounded as follows.
\paragraph{Term 1:} With the H\"{o}lder's inequality, we have
\begin{align*}
    & \bar h_{{k \otimes y}}\left|(\ve_{\tilde a} - \ve_{\tilde a'})^\top   \mLambda_{\tilde\vp_{y}}^{-1} \left(\mT_y^{-1}\mT_{{k \otimes y}}   - \mI \right)^\top\mLambda_{\tilde \vp_{y}} \bm 1\right|\\
\le & \bar h_{{k \otimes y}}\left\|\ve_{\tilde a} - \ve_{\tilde a'}\right\|_{1}   \left\|\mLambda_{\tilde\vp_{y}}^{-1} \left(\mT_y^{-1}\mT_{{k \otimes y}}   - \mI \right)^\top\mLambda_{\tilde \vp_{y}} \bm 1\right\|_\infty \\
\le & 2\bar h_{{k \otimes y}}  \left\|\mLambda_{\tilde\vp_{y}}^{-1} \left(\mT_y^{-1}\mT_{{k \otimes y}}   - \mI \right)^\top\mLambda_{\tilde \vp_{y}} \bm 1\right\|_\infty \\
\le & 2\bar h_{{k \otimes y}}  \left\|\mLambda_{\tilde\vp_{y}}^{-1} \left(\mT_y^{-1}\mT_{{k \otimes y}}   - \mI \right)^\top\mLambda_{\tilde \vp_{y}} \right\|_\infty \\
 =  & 2\bar h_{{k \otimes y}}  \left\| \mLambda_{\tilde \vp_{y}}\left(\mT_y^{-1}\mT_{{k \otimes y}}   - \mI \right)\mLambda_{\tilde\vp_{y}}^{-1} \right\|_1 \\
\end{align*}
\paragraph{Term 2:} 
Denote by $\delta_{k\otimes y} := \max\limits_{a\in[M]}~|H[a,k\otimes y] - \bar h_{k\otimes y}|$, which is the largest absolute offset from its mean.
With the H\"{o}lder's inequality, we have
\begin{align*}
    &  \left|(\ve_{\tilde a} - \ve_{\tilde a'})^\top  \left(\mLambda_{\tilde\vp_{y}}^{-1}\mT_{{k \otimes y}}^\top  \mLambda_{\vp_{y}} - \mI\right) \vv_{{k \otimes y}}\right|\\
\le & \|\ve_{\tilde a} - \ve_{\tilde a'}\|_{1}  \left\|\left(\mLambda_{\tilde\vp_{y}}^{-1}\mT_{{k \otimes y}}^\top  \mLambda_{\vp_{y}} - \mI\right) \vv_{{k \otimes y}}\right\|_{\infty}\\
\le & 2  \left\|\left(\mLambda_{\tilde\vp_{y}}^{-1}\mT_{{k \otimes y}}^\top  \mLambda_{\vp_{y}} - \mI\right) \vv_{{k \otimes y}}\right\|_{\infty} \\
\le & 2 \delta_{k\otimes y} \left\|\mLambda_{\tilde\vp_{y}}^{-1}\mT_{{k \otimes y}}^\top  \mLambda_{\vp_{y}} - \mI \right\|_{\infty} \\
 =  & 2 \delta_{k\otimes y} \left\| \mLambda_{\vp_{y}}\mT_{{k \otimes y}} \mLambda_{\tilde\vp_{y}}^{-1} - \mI \right\|_{1} \\
\end{align*}

\paragraph{Wrap-up:}
\begin{align*}
    & \left|~\left|(\ve_{\tilde a} - \ve_{\tilde a'})^\top \widetilde \mH[:{k \otimes y}]\right| - \left|(\ve_{\tilde a} - \ve_{\tilde a'})^\top  \mH[:{k \otimes y}]\right| ~ \right|\\
\le & 2\bar h_{{k \otimes y}}  \left\| \mLambda_{\tilde \vp_{y}}\left(\mT_y^{-1}\mT_{{k \otimes y}}   - \mI \right)\mLambda_{\tilde\vp_{y}}^{-1} \right\|_1 + 2 \delta_{k\otimes y} \left\| \mLambda_{\vp_{y}}\mT_{{k \otimes y}} \mLambda_{\tilde\vp_{y}}^{-1} - \mI \right\|_{1}.
\end{align*}
Denote by $\widetilde\Delta^{\tilde a,\tilde a'}_{{k \otimes y}}:= |\widetilde \mH[\tilde a, {k \otimes y}] -  \widetilde \mH[\tilde a', {k \otimes y}]|$ the noisy disparity and $\Delta^{\tilde a,\tilde a'}_{{k \otimes y}}:= | \mH[\tilde a, {k \otimes y}] -  \mH[\tilde a', {k \otimes y}]|$ the clean disparity between attributes $\tilde a$ and $\tilde a'$ in the case when $f(X)=k$ and $Y=y$.
We have
\begin{align*}
    & \left|\widetilde\Delta^{\EO}(\widetilde{\mathcal D}, f) - \Delta^{\EO}(\mathcal D, f)\right| \\
\le & \frac{1}{M(M-1)K^2}\sum_{\tilde a, \tilde a' \in [M], k,y\in[K]} \left|\widetilde\Delta^{\tilde a,\tilde a'}_{{k \otimes y}} - \Delta^{\tilde a,\tilde a'}_{{k \otimes y}}\right| \\
\le &  \frac{2}{M(M-1)K^2}\sum_{\tilde a, \tilde a' \in [M], k,y\in[K]} \left( \bar h_{{k \otimes y}}  \left\| \mLambda_{\tilde \vp_{y}}\left(\mT_y^{-1}\mT_{{k \otimes y}}   - \mI \right)\mLambda_{\tilde\vp_{y}}^{-1} \right\|_1 +  \delta_{k\otimes y} \left\| \mLambda_{\vp_{y}}\mT_{{k \otimes y}} \mLambda_{\tilde\vp_{y}}^{-1} - \mI \right\|_{1} \right) \\
= &  \frac{2}{K^2}\sum_{k,y\in[K]} \left( \bar h_{{k \otimes y}}  \left\| \mLambda_{\tilde \vp_{y}}\left(\mT_y^{-1}\mT_{{k \otimes y}}   - \mI \right)\mLambda_{\tilde\vp_{y}}^{-1} \right\|_1 +  \delta_{k\otimes y} \left\| \mLambda_{\vp_{y}}\mT_{{k \otimes y}} \mLambda_{\tilde\vp_{y}}^{-1} - \mI \right\|_{1} \right).
\end{align*}

The results for \DP\ can be obtained by dropping the dependence on $Y=y$, and the results for \EOp\ can be obtained by letting $k=1$ and $y=1$.
\end{proof}

\subsection{Full Version of Theorem~\ref{lem:relation_true_noisy} and Its Proof}\label{appendix:lemma_relation_true_noisy}

Recall $\vp$, $\tilde\vp$, $\mT$ and $\mT_k$ are clean prior, noisy prior, global transition matrix, and local transition matrix defined in Sec.~\ref{sec:pre}. Denote by $\bm\mLambda_{\tilde \vp}$ and $\bm\mLambda_{\vp}$ the square diagonal matrices constructed from $\tilde \vp$ and $\vp$. 

\textbf{Theorem 4.1} (Closed-form relationship (\DP,\EO,\EOp)).
The relationship between the true fairness vector $\vh^\uni$ and the corresponding noisy fairness vector $\tilde\vh^\uni$ writes as
\[
\vh^\uni = ({\mT^\uni}^\top\mLambda_{\vp^\uni})^{-1}\mLambda_{\tilde{\vp}^\uni}\tilde \vh^\uni, \quad \forall \uni \in\{\DP,\EO,\EOp\},
\]
where $\bm\mLambda_{\tilde \vp^\uni}$ and $\bm\mLambda_{\vp^\uni}$ denote the square diagonal matrix constructed from $\tilde \vp^\uni$ and $\vp^\uni$, $\uni$ unifies different fairness metrics. Particularly,
\squishlist
\item \DP~($\forall k\in[K]$): $\vp^{\DP}:=[\PP(A=1), \cdots, \PP(A=M) ]^\top$, $\tilde{\vp}^{\DP}:=[\PP(\widetilde A=1), \cdots, \PP(\widetilde A=M) ]^\top$. $\mT^{\DP} := \mT_k$, where the $(a,\tilde a)$-th element of $\mT_k$ is $T_{k}[{a,\tilde a}] := \PP( \widetilde A = \tilde a | f(X)=k , A = a)$.
\begin{align*}
  \vh^\DP:=\mH{[:,k]} & :=[ \PP(f(X) = k| A=1),  \cdots, \PP(f(X) = k| A=M)]^\top  \\
  \tilde\vh^\DP:=\widetilde \mH{[:,k]} & :=[ \PP(f(X) = k|\widetilde A=1),  \cdots, \PP(f(X) = k|\widetilde A=M)]^\top.
\end{align*}
\item \EO\ and \EOp\ ($\forall k,y\in[K], \uni \in\{\EO, \EOp\}$): $\forall k,y\in[K]$: $k\otimes y:=K(k-1)+y$, $\vp^{\uni}:=\vp_{y}:=[\PP(A=1|Y=y), \cdots, \PP(A=M|Y=y) ]^\top$, $\tilde\vp^{\uni}:=\tilde{\vp}_{y}:=[\PP(\widetilde A=1|Y=y), \cdots, \PP(\widetilde A=M|Y=y) ]^\top$.
$\mT^{\uni}:=\mT_{{k \otimes y}}$, where the $(a,\tilde a)$-th element of $\mT_{{k \otimes y}}$ is $T_{{k \otimes y}}[{a,\tilde a}] := \PP( \widetilde A = \tilde a | f(X)=k , Y=y, A = a)$.
\begin{align*}
  \vh^\uni:=\mH{[:,k\otimes y]} & :=[ \PP(f(X) = k|Y=y, A=1),  \cdots, \PP(f(X) = k|Y=y, A=M)]^\top  \\
  \tilde\vh^\uni:=\widetilde \mH{[:,k\otimes y]} & :=[ \PP(f(X) = k|Y=y,\widetilde A=1),  \cdots, \PP(f(X) = k|Y=y,\widetilde A=M)]^\top.
\end{align*}
\squishend
\begin{proof}

We first prove the theorem for \DP, then for \EO\ and \EOp.

\para{Proof for \DP.}
In \DP, each element of $\tilde\vh^\DP$ satisfies:
\begin{align*}
    & \PP(f(X)=k | \widetilde A = \tilde a) \\   
  = & \frac{  \sum_{a\in[M]} \PP(f(X)=k , \widetilde A = \tilde a, A = a)}{ \PP(\widetilde A = \tilde a)} \\   
  = & \frac{  \sum_{a\in[M]} \PP( \widetilde A = \tilde a | f(X)=k , A = a)  \cdot \PP(A=a) \cdot \PP(f(X)=k | A = a)}{ \PP(\widetilde A = \tilde a)}   
\end{align*}

Recall $\mT_{k}$ is the attribute noise transition matrix when $f(X)=k$, where the $(a,\tilde a)$-th element is $T_{k}[{a,\tilde a}] := \PP( \widetilde A = \tilde a | f(X)=k , A = a)$. Recall $\vp:=[\PP(A=1), \cdots, \PP(A=M) ]^\top$ and $\tilde{\vp}:=[\PP(\widetilde A=1), \cdots, \PP(\widetilde A=M) ]^\top$ the clean prior probabilities and noisy prior probability, respectively.
The above equation can be re-written as a matrix form as

\[
\widetilde \mH[:,k] = \mLambda_{\tilde{\vp}}^{-1} \mT_{k}^\top  \mLambda_{\vp}\mH[:,k],
\]
which is equivalent to
\[
 \mH[:,k] = ((\mT_{k}^\top)\mLambda_{\vp})^{-1}  \mLambda_{\tilde{\vp}}   \widetilde \mH[:,k].
\]

\para{Proof for \EO, \EOp.}
In \EO\ or \EOp, each element of $\tilde \vh^\uni$ satisfies:
\begin{align*}
   & \PP(f(X)=k|Y=y, \widetilde A = \tilde{a}) \\
 = & \frac{\PP(f(X)=k,Y=y, \widetilde A = \tilde{a})}{\PP(Y=y,\widetilde A=\tilde a)} \\
 = &\frac{  \sum_{a\in[M]} \PP(f(X)=k,Y=y, \widetilde A = \tilde{a}, A=a)}{\PP(Y=y,\widetilde A=\tilde a)} \\
 = &\frac{  \sum_{a\in[M]} \PP( \widetilde A = \tilde{a} | f(X)=k,Y=y,  A=a) \cdot \PP(Y=y,A=a) \cdot \PP(f(X)=k|Y=y,A=a) }{\PP(Y=y,\widetilde A=\tilde a)} 
\end{align*}

Denote by $\mT_{{k \otimes y}}$ the attribute noise transition matrix when $f(X)=k$ and $Y=y$, where the $(a,\tilde a)$-th element is $\mT_{{k \otimes y}}[{a,\tilde a}] := \PP( \widetilde A = \tilde a | f(X)=k , Y=y, A = a)$. Denote by $\vp_y:=[\PP(A=1|Y=y), \cdots, \PP(A=K|Y=y) ]^\top$ and $\tilde{\vp}_y:=[\PP(\widetilde A=1|Y=y), \cdots, \PP(\widetilde A=K|Y=y) ]^\top$ the clean prior probabilities and noisy prior probability, respectively.
The above equation can be re-written as a matrix form as
\[
  \widetilde\mH[:,k] = \mLambda_{\tilde{\vp}_y}^{-1} \mT_{{k \otimes y}}^\top \mLambda_{\vp_y}  \mH[:,k],
\]
which is equivalent to
\[
  \mH[:,k] = (\mT_{{k \otimes y}}^\top \mLambda_{\vp_y})^{-1} \mLambda_{\tilde{\vp}_y} \widetilde\mH[:,k].
\]

\para{Wrap-up.} We can conclude the proof by unifying the above two results with $\uni$.
\end{proof}

\subsection{Proof for Corollary~\ref{coro:exact_base_error}}

\begin{proof}

When the conditional independence (Assumption~\ref{def:tl-ind}) $$\PP(\widetilde A =a'|A=a,Y=y) = \PP(\widetilde A =a'|A=a,f(X)=k,Y=y), \forall a',a\in[M]$$ holds,
we have $\mT_y = \mT_{{k \otimes y}}$ and Term-1 in Theorem~\ref{thm:baseline_error} can be dropped.
For Term-2, to get a tight bound in this specific case, we apply the H\"{o}lder's inequality by using $l_\infty$ norm on $\ve_{\tilde a} - \ve_{\tilde a'}$, i.e.,
\begin{align*}
    &  \left|(\ve_{\tilde a} - \ve_{\tilde a'})^\top  \left(\mLambda_{\tilde\vp_{y}}^{-1}\mT_{{k \otimes y}}^\top  \mLambda_{\vp_{y}} - \mI\right) \vv_{{k \otimes y}}\right|\\
\le & \|\ve_{\tilde a} - \ve_{\tilde a'}\|_{\infty}  \left\|\left(\mLambda_{\tilde\vp_{y}}^{-1}\mT_{{k \otimes y}}^\top  \mLambda_{\vp_{y}} - \mI\right) \vv_{{k \otimes y}}\right\|_{1}\\
 =  &  \left\|\left(\mLambda_{\tilde\vp_{y}}^{-1}\mT_{{k \otimes y}}^\top  \mLambda_{\vp_{y}} - \mI\right) \vv_{{k \otimes y}}\right\|_{1} \\
\le & K \cdot \delta_{k\otimes y} \left\|\mLambda_{\tilde\vp_{y}}^{-1}\mT_{{k \otimes y}}^\top  \mLambda_{\vp_{y}} - \mI \right\|_{1} \\
 =  & K \cdot \delta_{k\otimes y} \left\| \mLambda_{\vp_{y}}\mT_{{k \otimes y}} \mLambda_{\tilde\vp_{y}}^{-1} - \mI \right\|_{\infty} \\
\end{align*}
Therefore,
\begin{align*}
    & \left|\widetilde\Delta^{\EO}(\widetilde{\mathcal D}, f) - \Delta^{\EO}(\mathcal D, f)\right| \\
\le &  \frac{1}{K}\sum_{k,y\in[K]}   \delta_{k\otimes y} \left\| \mLambda_{\vp_{y}}\mT_{{k \otimes y}} \mLambda_{\tilde\vp_{y}}^{-1} - \mI \right\|_{\infty} \\
 =  &  \frac{1}{K}\sum_{k,y\in[K]}   \delta_{k\otimes y} \left\| \mLambda_{\vp_{y}}\mT_{{y}} \mLambda_{\tilde\vp_{y}}^{-1} - \mI \right\|_{\infty} \\
 =  &  \frac{1}{K}\sum_{k,y\in[K]}   \delta_{k\otimes y} \left\| \check\mT_{y} - \mI \right\|_{\infty},
\end{align*}
where $\check T_{y}[a,\tilde a] = \PP( A=a | \widetilde A=\tilde a, Y=y)$.

\paragraph{Special binary case in \DP}
In addition to the conditional independence, when the sensitive attribute is binary and the label class is binary, considering \DP, we have
\begin{align*}
    & \left|\widetilde\Delta^{\DP}(\widetilde{\mathcal D}, f) - \Delta^{\DP}(\mathcal D, f)\right| 
\le  2   \delta_{k} \left\| \check\mT - \mI \right\|_{\infty},
\end{align*}
where $\check T_{y}[a,\tilde a] = \PP( A=a | \widetilde A=\tilde a)$.
Let $\check T_{y}[1,2] = e_1, \check T_{y}[2,1] = e_2$, we know
\begin{align*}
    \check\mT := 
    \begin{pmatrix}
    1-e_2 & e_1 \\
    e_2 & 1 - e_1
    \end{pmatrix}
\end{align*}
and
\begin{align*}
    &  \left|\widetilde\Delta^{\DP}(\widetilde{\mathcal D}, f) - \Delta^{\DP}(\mathcal D, f)\right| \le  2{\delta_k} \cdot  (e_1 + e_2).
\end{align*}
Note the equality in above inequality always holds. To prove it, firstly we note
\begin{align*}
    & \PP(f(X)=k | \widetilde A = \tilde a) \\   
  = & \frac{  \sum_{a\in[M]} \PP(f(X)=k , \widetilde A = \tilde a, A = a)}{ \PP(\widetilde A = \tilde a)} \\   
  = & \frac{  \sum_{a\in[M]} \PP( \widetilde A = \tilde a | f(X)=k , A = a)  \cdot \PP(A=a) \cdot \PP(f(X)=k | A = a)}{ \PP(\widetilde A = \tilde a)}\\  
  = & \frac{  \sum_{a\in[M]} \PP( \widetilde A = \tilde a | A = a)  \cdot \PP(A=a) \cdot \PP(f(X)=k | A = a)}{ \PP(\widetilde A = \tilde a)}  \\
  = &  \sum_{a\in[M]} \PP(A = a |  \widetilde A = \tilde a)   \cdot \PP(f(X)=k | A = a),
\end{align*}
\ie $ \widetilde \mH[:,{k}] = \check \mT^\top \mH[:,{k}].$
Denote by $\mH[:,{1}]=[h,h']^\top$.
We have ($\tilde a\ne \tilde a'$)
\[
\left|(\ve_{\tilde a} - \ve_{\tilde a'})^\top \widetilde \mH[:,{1}]\right| = |h-h'| \cdot |1-e_1-e_2|,
\]
and 
\[
\left|(\ve_{\tilde a} - \ve_{\tilde a'})^\top \mH[:,{1}]\right| = |h-h'|.
\]
Therefore, letting $\tilde a = 1, \tilde a =2$, we have
\begin{align*}
   & \left|\widetilde\Delta^{\DP}(\widetilde{\mathcal D}, f) - \Delta^{\DP}(\mathcal D, f)\right| \\
 = & \frac{1}{2}\sum_{k\in\{1,2\}}\left|~\left|(\ve_{1} - \ve_{2})^\top \widetilde \mH[:,{k}]\right| - \left|(\ve_{1} - \ve_{2})^\top  \mH[:,{k}]\right| ~ \right| \\
 = & \left|~\left|(\ve_{1} - \ve_{2})^\top \widetilde \mH[:,{1}]\right| - \left|(\ve_{1} - \ve_{2})^\top  \mH[:,{1}]\right| ~ \right| \\
 = & |h-h'| \cdot |e_1+e_2|\\
 = & \delta \cdot (e_1 + e_2),
\end{align*}
where $\delta = |\PP(f(X)=1|A=1) - \PP(f(X)=1|A=2)|$.
Therefore, the equality holds.

\end{proof}

\subsection{Proof for Theorem~\ref{thm:error_correction}} 

\textbf{Theorem 4.5} (Error upper bound of calibrated metrics).
Denote the error of the calibrated fairness metrics by $\ecal_\uni:= |\widehat\Delta^{\uni}(\widetilde{\mathcal D}, f) - \Delta^{\uni}(\mathcal D, f)|$. It can be upper bounded as:
\squishlist
\item \DP: 
\begin{align*}
\ecal_{\DP}
\le  \frac{2}{K}\sum_{k\in[K]} \left\|\mLambda_{{\vp}}^{-1} \right\|_1  \left\|\mLambda_{{\vp}} \mH[:,k]  \right\|_\infty \varepsilon(\widehat\mT_{k},\hat \vp),
\end{align*}
where $\varepsilon(\widehat\mT_k,\hat \vp) := \| \mLambda_{{\hat\vp}}^{-1}\mLambda_{{\vp}} - \mI\|_{1}\|\mT_{k}\widehat\mT_k^{-1}\|_{1} + \|  \mI -  \mT_{k}\widehat\mT_k^{-1}  \|_{1}$ is the error induced by calibration.
\item \EO: 
\begin{align*}
\ecal_\EO
\le  \frac{2}{K^2}\sum_{k\in[K],y\in[K]} \left\|\mLambda_{{\vp_y}}^{-1} \right\|_1  \left\|\mLambda_{{\vp_y}} \mH[:,k\otimes y]  \right\|_\infty \varepsilon(\widehat\mT_{k\otimes y},\hat \vp_y),
\end{align*}
where $\varepsilon(\widehat\mT_{k\otimes y},\hat \vp_y) := \| \mLambda_{{\hat\vp_y}}^{-1}\mLambda_{{\vp_y}} - \mI\|_{1}\|\mT_{k\otimes y}\widehat\mT_{k\otimes y}^{-1}\|_{1} + \|  \mI -  \mT_{k\otimes y}\widehat\mT_{k\otimes y}^{-1}  \|_{1}$ is the error induced by calibration.
\item \EOp: 
\begin{align*}
\ecal_\EOp
\le  2\sum_{k=1,y=1} \left\|\mLambda_{{\vp_y}}^{-1} \right\|_1  \left\|\mLambda_{{\vp_y}} \mH[:,k\otimes y]  \right\|_\infty \varepsilon(\widehat\mT_{k\otimes y},\hat \vp_y),
\end{align*}
where $\varepsilon(\widehat\mT_{k\otimes y},\hat \vp_y) := \| \mLambda_{{\hat\vp_y}}^{-1}\mLambda_{{\vp_y}} - \mI\|_{1}\|\mT_{k\otimes y}\widehat\mT_{k\otimes y}^{-1}\|_{1} + \|  \mI -  \mT_{k\otimes y}\widehat\mT_{k\otimes y}^{-1}  \|_{1}$ is the error induced by calibration.
\squishend

\begin{proof}

We prove with \EO. 

Consider the case when $f(X)=k$ and $Y=y$. For ease of notations, we use $\widehat \mT$ to denote the estimated local transition matrix (should be $\widehat \mT_{k\otimes y}$). Denote the noisy (clean) fairness vectors with respect to $f(X)=k$ and $Y=y$ by $\tilde \vh$ ($\vh$).
The error can be decomposed by
\begin{align*}
    &\Bigg| \left| (\ve_a - \ve_{a'})^\top \left(\mLambda_{{\hat\vp}_{y}}^{-1} (\widehat\mT^\top)^{-1} \mLambda_{\tilde{\vp}_{y}} \widetilde \vh \right) \right| -  \left| (\ve_a - \ve_{a'})^\top \left(\mLambda_{{\vp}_{y}}^{-1} (\mT_{{k \otimes y}}^\top)^{-1} \mLambda_{\tilde{\vp}_{y}} \widetilde \vh \right) \right| \Bigg| \\
=   & \underbrace{\left| (\ve_a - \ve_{a'})^\top \left( (\mLambda_{{\hat\vp}_{y}}^{-1} - \mLambda_{{\vp}_{y}}^{-1}) (\widehat\mT^\top)^{-1} \mLambda_{\tilde{\vp}_{y}} \widetilde \vh \right) \right|}_{\text{Term-1}} \\
    & + \underbrace{\Bigg| \left| (\ve_a - \ve_{a'})^\top \left(\mLambda_{{\vp}_{y}}^{-1} (\widehat\mT^\top)^{-1} \mLambda_{\tilde{\vp}_{y}} \widetilde \vh \right) \right| -  \left| (\ve_a - \ve_{a'})^\top \left(\mLambda_{{\vp}_{y}}^{-1} (\mT_{{k \otimes y}}^\top)^{-1} \mLambda_{\tilde{\vp}_{y}} \widetilde \vh \right) \right| \Bigg|}_{\text{Term-2}}.
\end{align*}

Now we upper bound them respectively.

\para{Term-1:}
\begin{align*}
    & \left| (\ve_a - \ve_{a'})^\top \left( (\mLambda_{{\hat\vp}_{y}}^{-1} - \mLambda_{{\vp}_{y}}^{-1}) (\widehat\mT^\top)^{-1} \mLambda_{\tilde{\vp}_{y}} \widetilde \vh \right) \right| \\
\overset{(a)}{=} 
    & \left| (\ve_a - \ve_{a'})^\top \left( (\mLambda_{{\hat\vp}_{y}}^{-1} - \mLambda_{{\vp}_{y}}^{-1}) (\mT_{{k \otimes y}}\widehat\mT^{-1})^\top \mLambda_{\vp_{y}} \mH[:,{k \otimes y}] \right) \right| \\
\overset{(b)}{=} 
    & \left| (\ve_a - \ve_{a'})^\top \left( (\mLambda_{{\hat\vp}_{y}}^{-1}\mLambda_{{\vp}_{y}} - \mI) \mLambda_{{\vp}_{y}}^{-1} \mT_{\delta}^\top \mLambda_{\vp_{y}} \mH[:,{k \otimes y}] \right) \right| \\
\le & 2\left\| \mLambda_{{\hat\vp}_{y}}^{-1}\mLambda_{{\vp}_{y}} - \mI) \right\|_{\infty} \left\|\mLambda_{{\vp}_{y}}^{-1}\right\|_{\infty} \left\|\mT_{\delta}\right\|_{1} \left\|\mLambda_{\vp_{y}} \mH[:,{k \otimes y}] \right\|_{\infty}  \\
=   &  2 \left\|\mLambda_{{\vp}_{y}}^{-1} \right\|_\infty  \left\|\mLambda_{{\vp}_{y}} \mH[:,{k \otimes y}]  \right\|_\infty \left(\left\| \mLambda_{{\hat\vp}_{y}}^{-1}\mLambda_{{\vp}_{y}} - \mI) \right\|_{\infty}\left\|\mT_{\delta}\right\|_{1}\right),
\end{align*}
where equality $(a)$ holds due to
\begin{align*}
    &  \mLambda_{\tilde{\vp}_{y}}\widetilde \mH[:,{k \otimes y}] = \mT_{{k \otimes y}}^\top \mLambda_{\vp_{y}} \mH[:,{k \otimes y}]
\end{align*}
and equality $(b)$ holds because we denote the error matrix by $\mT_{\delta}$, \ie
$$\widehat\mT =\mT_\delta^{-1} \mT_{{k \otimes y}} \Leftrightarrow \mT_\delta = \mT_{{k \otimes y}}\widehat\mT^{-1}. $$

\para{Term-2:}
Before preceeding, we introduce the Woodbury matrix identity:
\[
(\mA+\mU\mC\mV)^{-1} = \mA^{-1} - \mA^{-1} \mU (\mC^{-1} + \mV\mA^{-1} \mU)^{-1} \mV \mA^{-1}
\]
Let $A:=\mT_{{k \otimes y}}^\top$, $\mC=\mI$, $\mV:=\mI$, $\mU:=\widehat\mT^\top - \mT_{{k \otimes y}}^\top$. By Woodbury matrix identity, we have
\begin{align*}
  & (\widehat\mT^\top)^{-1} \\
= & (\widehat\mT_{{k \otimes y}}^\top+ (\widehat\mT^\top - \mT_{{k \otimes y}}^\top))^{-1} \\
= & (\mT_{{k \otimes y}}^\top)^{-1} - (\mT_{{k \otimes y}}^\top)^{-1} (\widehat\mT^\top - \mT_{{k \otimes y}}^\top) \left(\mI + (\mT_{{k \otimes y}}^\top)^{-1} (\widehat\mT^\top - \mT_{{k \otimes y}}^\top)\right)^{-1} (\mT_{{k \otimes y}}^\top)^{-1}
\end{align*}

Term-2 can be upper bounded as:
{\small\begin{align*}
    &\Bigg| \left| (\ve_a - \ve_{a'})^\top \left(\mLambda_{\vp_{y}}^{-1} (\widehat\mT^\top)^{-1} \mLambda_{\tilde\vp_{y}} \widetilde \vh \right) \right| -  \left| (\ve_a - \ve_{a'})^\top \left(\mLambda_{\vp_{y}}^{-1} (\mT_{{k \otimes y}}^\top)^{-1} \mLambda_{\tilde\vp_{y}} \widetilde \vh \right) \right| \Bigg| \\
\overset{(a)}{=}   & \Bigg| \left| (\ve_a - \ve_{a'})^\top \left(\mLambda_{\vp_{y}}^{-1} \left( (\mT_{{k \otimes y}}^\top)^{-1} - (\mT_{{k \otimes y}}^\top)^{-1} (\widehat\mT^\top - \mT_{{k \otimes y}}^\top) \left(\mI + (\mT_{{k \otimes y}}^\top)^{-1} (\widehat\mT^\top - \mT_{{k \otimes y}}^\top)\right)^{-1} (\mT_{{k \otimes y}}^\top)^{-1} \right) \mLambda_{\tilde\vp_{y}} \widetilde \vh \right) \right| \\
    & -  \left| (\ve_a - \ve_{a'})^\top \left(\mLambda_{\vp_{y}}^{-1} (\mT_{{k \otimes y}}^\top)^{-1} \mLambda_{\tilde\vp_{y}} \widetilde \vh \right) \right| \Bigg| \\
\le &  \left| (\ve_a - \ve_{a'})^\top \left(\mLambda_{\vp_{y}}^{-1} (\mT_{{k \otimes y}}^\top)^{-1} (\widehat\mT^\top - \mT_{{k \otimes y}}^\top) \left(\mI + (\mT_{{k \otimes y}}^\top)^{-1} (\widehat\mT^\top - \mT_{{k \otimes y}}^\top)\right)^{-1} (\mT_{{k \otimes y}}^\top)^{-1} \mLambda_{\tilde\vp_{y}} \widetilde \vh \right) \right| \\
\overset{(b)}{\le} &  \| \ve_a - \ve_{a'} \|_1 \left\|\mLambda_{\vp_{y}}^{-1} (\mT_{{k \otimes y}}^\top)^{-1} (\widehat\mT^\top - \mT_{{k \otimes y}}^\top) \left(\mI + (\mT_{{k \otimes y}}^\top)^{-1} (\widehat\mT^\top - \mT_{{k \otimes y}}^\top)\right)^{-1} (\mT_{{k \otimes y}}^\top)^{-1} \mLambda_{\tilde\vp_{y}} \widetilde \vh \right\|_\infty \\
\le &  2 \left\|\mLambda_{\vp_{y}}^{-1} \right\|_\infty \left\| (\mT_{{k \otimes y}}^\top)^{-1} (\widehat\mT^\top - \mT_{{k \otimes y}}^\top) \left(\mI + (\mT_{{k \otimes y}}^\top)^{-1} (\widehat\mT^\top - \mT_{{k \otimes y}}^\top)\right)^{-1} (\mT_{{k \otimes y}}^\top)^{-1} \mLambda_{\tilde\vp_{y}} \widetilde \vh \right\|_\infty \\
= &  2 \left\|\mLambda_{\vp_{y}}^{-1} \right\|_\infty \left\| \left(\mI + (\mT_{{k \otimes y}}^\top)^{-1} (\widehat\mT^\top - \mT_{{k \otimes y}}^\top) - \mI\right) \left(\mI + (\mT_{{k \otimes y}}^\top)^{-1} (\widehat\mT^\top - \mT_{{k \otimes y}}^\top)\right)^{-1} (\mT_{{k \otimes y}}^\top)^{-1} \mLambda_{\tilde\vp_{y}} \widetilde \vh \right\|_\infty \\
= &  2 \left\|\mLambda_{\vp_{y}}^{-1} \right\|_\infty \left\| \left[ \mI - \left(\mI + (\mT_{{k \otimes y}}^\top)^{-1} (\widehat\mT^\top - \mT_{{k \otimes y}}^\top)\right)^{-1} \right](\mT_{{k \otimes y}}^\top)^{-1} \mLambda_{\tilde\vp_{y}} \widetilde \vh \right\|_\infty \\
= &  2 \left\|\mLambda_{\vp_{y}}^{-1} \right\|_\infty \left\| \left( \mI -  \mT_{{k \otimes y}} \widehat\mT^{-1}   \right)^\top(\mT_{{k \otimes y}}^\top)^{-1} \mLambda_{\tilde\vp_{y}} \widetilde \vh \right\|_\infty \\
\overset{(c)}{\le} &  2 \left\|\mLambda_{\vp_{y}}^{-1} \right\|_\infty \left\|  \mI -  \mT_{\delta}   \right\|_{1} \left\|(\mT_{{k \otimes y}}^\top)^{-1} \mLambda_{\tilde\vp_{y}} \widetilde \vh \right\|_\infty \\
\overset{(d)}{=} &  2 \left\|\mLambda_{\vp_{y}}^{-1} \right\|_\infty \left\|  \mI -  \mT_{\delta}   \right\|_{1} \left\|\mLambda_{\vp_{y}} \mH[:,{k \otimes y}]  \right\|_\infty,
\end{align*}
}
where the key steps are:
\squishlist
\item (a): Woodbury identity. 
\item (b): H\"{o}lder's inequality.
\item (c): $\widehat\mT =\mT_\delta^{-1} \mT_{{k \otimes y}} $ and triangle inequality
\item (d):
\begin{align*}
    & \widetilde \mH[:,{k \otimes y}] = \mLambda_{\tilde\vp_{y}}^{-1} \mT_{{k \otimes y}}^\top \mLambda_{\vp_{y}} \mH[:,{k \otimes y}] \\
\Leftrightarrow & (\mT_{{k \otimes y}}^\top)^{-1}\mLambda_{\tilde\vp_{y}} \widetilde \mH[:,{k \otimes y}] =   \mLambda_{\vp_{y}} \mH[:,{k \otimes y}].
\end{align*}
\squishend

\para{Wrap-up}
Combining the upper bounds of Term-1 and Term-2, we have (recovering full notations)
\begin{align*}
    &\Bigg| \left| (\ve_a - \ve_{a'})^\top \left(\mLambda_{{\hat\vp}_{y}}^{-1} (\widehat\mT^\top)^{-1} \mLambda_{\tilde{\vp}_{y}} \widetilde \vh \right) \right| -  \left| (\ve_a - \ve_{a'})^\top \left(\mLambda_{{\vp}_{y}}^{-1} (\mT_{{k \otimes y}}^\top)^{-1} \mLambda_{\tilde{\vp}_{y}} \widetilde \vh \right) \right| \Bigg| \\
\le &  2 \left\|\mLambda_{{\vp}_{y}}^{-1} \right\|_\infty  \left\|\mLambda_{{\vp}_{y}} \mH[:,{k \otimes y}]  \right\|_\infty \left(\left\| \mLambda_{{\hat\vp}_{y}}^{-1}\mLambda_{{\vp}_{y}} - \mI) \right\|_{\infty}\left\|\mT_{\delta}\right\|_{1} + \left\|  \mI -  \mT_{\delta}   \right\|_{1}\right) \\
 =  &  2 \left\|\mLambda_{{\vp}_{y}}^{-1} \right\|_\infty  \left\|\mLambda_{{\vp}_{y}} \mH[:,{k \otimes y}]  \right\|_\infty \left(\left\| \mLambda_{{\hat\vp}_{y}}^{-1}\mLambda_{{\vp}_{y}} - \mI) \right\|_{\infty}\left\|\mT_{{k \otimes y}}\widehat\mT_{k\otimes y}^{-1}\right\|_{1} + \left\|  \mI -  \mT_{{k \otimes y}}\widehat\mT_{k\otimes y}^{-1}   \right\|_{1}\right).
\end{align*}

Denote by $\widehat\Delta^{\tilde a,\tilde a'}_{{k \otimes y}}:= |\widehat \mH[\tilde a, {k \otimes y}] -  \widehat \mH[\tilde a', {k \otimes y}]|$ the calibrated disparity and $\Delta^{\tilde a,\tilde a'}_{{k \otimes y}}:= | \mH[\tilde a, {k \otimes y}] -  \mH[\tilde a', {k \otimes y}]|$ the clean disparity between attributes $\tilde a$ and $\tilde a'$ in the case when $f(X)=k$ and $Y=y$.
We have
\begin{align*}
    & \left|\widehat\Delta^{\EO}(\widetilde{\mathcal D}, f) - \Delta^{\EO}(\mathcal D, f)\right| \\
\le & \frac{1}{M(M-1)K^2}\sum_{\tilde a, \tilde a' \in [M], k,y\in[K]} \left|\widehat\Delta^{\tilde a,\tilde a'}_{{k \otimes y}} - \Delta^{\tilde a,\tilde a'}_{{k \otimes y}}\right| \\
\le &  \frac{2}{K^2}\sum_{k,y\in[K]} 2 \left\|\mLambda_{{\vp}_{y}}^{-1} \right\|_\infty  \left\|\mLambda_{{\vp}_{y}} \mH[:,{k \otimes y}]  \right\|_\infty \left(\left\| \mLambda_{{\hat\vp}_{y}}^{-1}\mLambda_{{\vp}_{y}} - \mI) \right\|_{\infty}\left\|\mT_{{k \otimes y}}\widehat\mT_{k\otimes y}^{-1}\right\|_{1} + \left\|  \mI -  \mT_{{k \otimes y}}\widehat\mT_{k\otimes y}^{-1}   \right\|_{1}\right).
\end{align*}

The above inequality can be generalized to \DP\ by dropping dependency on $y$ and to \EOp\ by requiring $k=1$ and $y=1$.

\end{proof}

\subsection{Proof for Corollary~\ref{coro:compare}}
\begin{proof}
Consider \DP.
Denote by $\mH[:,k=1]=[h,h']^\top$. We know $\delta = |h-h'|/2 = \Delta^\DP(\mathcal D,f)/2$.
Suppose $p \le 1/2$, $\left\|\mLambda_{{\vp}}^{-1} \right\|_\infty = 1/p$ and 
$$\left\|\mLambda_{{\vp}} \mH[:,k]  \right\|_\infty = \max(ph,(1-p)h').$$
Recall $$\varepsilon(\widehat\mT_k,\hat \vp) := \| \mLambda_{{\hat\vp}}^{-1}\mLambda_{{\vp}} - \mI\|_{1}\|\mT_{k}\widehat\mT_k^{-1}\|_{1} + \|  \mI -  \mT_{k}\widehat\mT_k^{-1}  \|_{1}.$$
By requiring the error upper bound in Theorem~\ref{thm:error_correction} less than the exact error in Corollary~\ref{coro:exact_base_error}, we have (when $k=1$)
\begin{align*}
    & \left\|\mLambda_{{\vp}}^{-1} \right\|_\infty  \left\|\mLambda_{{\vp}} \mH[:,k]  \right\|_\infty \varepsilon(\widehat\mT_k,\hat \vp) \le {\delta} \cdot  (e_1 + e_2) \\
\Leftrightarrow & \varepsilon(\widehat\mT_k,\hat \vp) \le \frac{{\delta} \cdot  (e_1 + e_2)}{\left\|\mLambda_{{\vp}}^{-1} \right\|_\infty  \left\|\mLambda_{{\vp}} \mH[:,k]  \right\|_\infty} \\
\Leftrightarrow & \varepsilon(\widehat\mT_k,\hat \vp) \le \frac{{\delta} \cdot  (e_1 + e_2)}{ \max(h,(1-p)h'/p)}.
\end{align*}
If $p=1/2$, noting $\max(h,h') = (|h+h'| + |h-h'|) / 2$, we further have (when $k=1$)
\begin{align*}
    \varepsilon(\widehat\mT_k,\hat \vp) \le \frac{ |h-h'| \cdot  (e_1 + e_2)}{ |h-h'| + |h+h'|} = \frac{  e_1 + e_2}{ 1 + \frac{h+h'}{|h-h'|}}  = \frac{  e_1 + e_2}{ 1 + \frac{h+h'}{\Delta^{\DP}(\mathcal D,f)}}.
\end{align*}

To make the above equality holds for all $k\in\{1,2\}$, we have
\begin{align*}
    \varepsilon(\widehat\mT_k,\hat \vp) \le \max_{k'\in\{1,2\}} \frac{  e_1 + e_2}{ 1 + \frac{\|\mH[:,k']\|_1}{\Delta^{\DP}(\mathcal D,f)}}, \forall k\in\{1,2\}.
\end{align*}

\end{proof}

\subsection{Differential Privacy Guarantee}\label{appendix:labeldp}

We explain how we calculate the differential privacy guarantee.

Suppose $\PP(\widetilde A = a | A = a, X) \le 1 - \epsilon_0$ and $\PP(\widetilde A = a | A = a', X) \ge  \epsilon_1, \forall X, a\in[M], a'\in[M], a\ne a' $. Then following the result of \citet{ghazi2021deep},
we have
\begin{align*}
\frac{\PP(\textsf{RandResponse}(a) = \tilde a)}{\PP(\textsf{RandResponse}(a') = \tilde a)} \le 
    \frac{\PP(\widetilde A = \tilde a | A = a, X)}{ \PP(\widetilde A = \tilde a | A = a', X)} \le 
    \frac{\max~ \PP(\widetilde A =   a | A = a, X)}{\min~ \PP(\widetilde A =  a | A = a', X)} \le \frac{1-\epsilon_0}{\epsilon_1} = e^\varepsilon.
\end{align*}
Then we know $\varepsilon = \ln(\frac{1-\epsilon_0}{\epsilon_1})$.
In practice, if proxies are too strong, \ie $\ln(\frac{1-\epsilon_0}{\epsilon_1})$ is too large, we can add additional noise to reduce their informativeness and therefore better protect privacy. 
For example, in experiments of Table~\ref{table:celebA_noise_facenet512}, when we add $40\%$ of random noise and reduce the proxy model accuracy to $58.45\%$, the the corresponding privacy guarantee is at least $0.41$-DP.
To get this value, noting the proxy model's accuracy of individual feature is not clear, we consider a native worst case that the model has an accuracy of $1$ on some feature. Then by adding $40\%$ of the random noise (random response), we have
\[
\epsilon = \ln\frac{1-0.4}{0.4} < 0.41,
\]
corresponding to at least $0.41$-DP.

\newpage

\section{More Discussions on Transition Matrix Estimators}\label{sec:discuss_T}

In this section, we first introduce how we adapt HOC to design our estimator for key statistics (Appendix~\ref{appendix:hoc_statestimator}), then extend it to a general form which can be used for \EO\ and \EOp\ (Appendix~\ref{appendix:hoc_general}).
For readers who are interested in details about HOC, we provide more details in Appendix~\ref{sec:hoc_detail}. We also encourage the readers to read the original papers \citep{zhu2021clusterability,zhu2022beyond}. For other possible estimators, we briefly discuss them in Appendix~\ref{appendix:estimator_training}.

\subsection{Adapting HOC}\label{appendix:hoc_statestimator}

Algorithm~\ref{alg:alg_hoc} shows how we adapt HOC as \texttt{StatEstimator} (in Algorithm~\ref{alg:alg_calibrate}, Line~5), namely \texttt{HOCFair}. The original HOC uses one proxy model and simulates the other two based on clusterability condition~\citep{zhu2021clusterability}, which assumes $x_n$ and its $2$-nearest-neighbors share the same true sensitive attribute, and therefore their noisy attributes can be used to simulate the output of proxy models.
If this condition does not hold~\citep{zhu2022beyond}, 
we can directly use more proxy models. With a sufficient number of noisy attributes, we can randomly select a subset of them for every sample as Line~7, and then approximate $\mT_k$ with $\widehat{\mT}_k$ in Line~\ref{alg_line:localT}. In our experiments, we test both using one proxy model and multiple proxy models.

\begin{algorithm}[!t]
\caption{StatEstimator: HOCFair (\DP)}\label{alg:alg_hoc} 
\begin{algorithmic}[1]
\STATE \textbf{Input:} Noisy dataset $\widetilde D$. Target model $f$. 
\\
\algcom{\# Get the number of noisy attributes (\ie \# proxy models)}
\STATE $C \leftarrow \texttt{\#Attribute}(\widetilde D)$ \\
\algcom{\# Get $2$-Nearest-Neighbors of $x_n$ and save their attributes as $x_n$'s attribute}
\IF{$C < 3$} %
\STATE { $ \{(x_n,y_n,(\tilde a_n^1, \cdots, \tilde a_n^{3C})) | n\in[N]\} \hspace{-3pt}\leftarrow \hspace{-3pt} \texttt{Get2NN}(\widetilde D)$}
\STATE { $\widetilde D  \leftarrow  \{(x_n,y_n,(\tilde a_n^1, \cdots, \tilde a_n^{3C})) | n\in[N]\} $}
\ENDIF
\\
 \algcom{\# Randomly sample 3 noisy attributes for each instance}
\STATE $\{(\tilde a_n^1, \tilde a_n^2, \tilde a_n^3) | n\in[N]\} \leftarrow \texttt{Sample}(\widetilde D)$  %
\\
\algcom{\# Get estimates $\vp\approx \hat\vp$}
\STATE $(\widehat\mT, \hat\vp) \leftarrow \textsf{HOC}(\{(\tilde a_n^1, \tilde a_n^2, \tilde a_n^3) | n\in[N]\})$ %
\\
\algcom{\# Get estimates $\mT_k \approx \widehat{\mT}_k$}
\STATE $(\widehat\mT_k, - ) \leftarrow \textsf{HOC}(\{(\tilde a_n^1, \tilde a_n^2, \tilde a_n^3) | n\in[N], f(x_n) = k\}), ~~\forall k\in[K]$ \label{alg_line:localT}  
\\
\algcom{ \# Return the estimated statistics}
\STATE \textbf{Output:} $\{\widehat\mT_1, \cdots, \widehat\mT_K\}, \hat\vp$
\end{algorithmic}
\end{algorithm}

\subsection{HOCFair: A General Form}\label{appendix:hoc_general}

Due to space limit, we only introduced the HOCFair specially designed for \DP (only depending on $f(X)$) in the main paper. Now we consider a general fairness metric depending on both $f(X)$ and $Y$. According to the full Version of Theorem~\ref{lem:relation_true_noisy} in Appendix~\ref{appendix:lemma_relation_true_noisy}, we need to estimate $\bm T_{k \otimes y}$ and $\bm p_y$, $\forall k \in [K], y\in[K]$.
We summarize the general form of HOCFair in Algorithm~\ref{alg:alg_hoc_general}. In this general case, our \globalT\ method in experiments adopt $\mT_{k\otimes y} \approx \widehat \mT$ and $\vp_y \approx \hat \vp_y, \forall y\in[K]$. For example, considering \EOp\ with binary attributes and binary label classes, we will estimate $4$ noise transition matrices and $2$ clean prior probabilities for \localT, and $1$ noise transition and $2$ clean prior probabilities for \globalT.

\begin{algorithm}[!h]
\caption{StatEstimator: HOCFair (General)}\label{alg:alg_hoc_general} 
\begin{algorithmic}[1]
\STATE \textbf{Input:} Noisy dataset $\widetilde D$. Target model $f$. 
\\
\algcom{\# Get the number of noisy attributes (\ie \# proxy models)}
\STATE $C \leftarrow \texttt{\#Attribute}(\widetilde D)$ \\
\algcom{\# Get $2$-Nearest-Neighbors of $x_n$ and save their attributes as $x_n$'s attribute}
\IF{$C < 3$} %
\STATE { $ \{(x_n,y_n,(\tilde a_n^1, \cdots, \tilde a_n^{3C})) | n\in[N]\} \hspace{-3pt}\leftarrow \hspace{-3pt} \texttt{Get2NN}(\widetilde D)$}
\STATE { $\widetilde D  \leftarrow  \{(x_n,y_n,(\tilde a_n^1, \cdots, \tilde a_n^{3C})) | n\in[N]\} $}
\ENDIF
\\
 \algcom{\# Randomly sample 3 noisy attributes for each instance}
\STATE $\{(\tilde a_n^1, \tilde a_n^2, \tilde a_n^3) | n\in[N]\} \leftarrow \texttt{Sample}(\widetilde D)$  %
\\
\algcom{\# Get estimates $\mT_k \approx \widehat{\mT}$ and $\vp\approx \hat\vp$}
\STATE $(\widehat\mT, \hat\vp) \leftarrow \textsf{HOC}(\{(\tilde a_n^1, \tilde a_n^2, \tilde a_n^3) | n\in[N]\})$ %
\\
\algcom{\# Get estimates $\mT_{k\otimes y} \approx \widehat{\mT}_{k\otimes y}$, and $\bm p_y = \hat {\bm p}_y$}
\FOR{$y \in [K]$}
\STATE $(\widehat\mT_{k\otimes y}, \hat {\bm p}_y ) \leftarrow \textsf{HOC}(\{(\tilde a_n^1, \tilde a_n^2, \tilde a_n^3) | n\in[N], f(x_n) = k, Y=y\}), ~~\forall k\in[K]$ 
\ENDFOR
\\
\algcom{ \# Return the estimated statistics}
\STATE \textbf{Output:} $\widehat\mT$, $\{\widehat\mT_{k\otimes y} ~|~ k\in[K], y\in[K]\}, \{\hat\vp_y ~|~ y\in[K]\}$
\end{algorithmic}
\end{algorithm}

\subsection{HOC}\label{sec:hoc_detail}
HOC \citep{zhu2021clusterability} relies on checking the agreements and disagreements among three noisy attributes of one feature. For example, given a three-tuple $(\tilde a_n^1, \tilde a_n^2, \tilde a_n^3)$, each noisy attribute may agree or disagree with the others. This consensus pattern encodes the information of noise transition matrix $\mT$. Suppose $(\tilde a_n^1, \tilde a_n^2, \tilde a_n^3)$ are drawn from random variables $(\tilde A^1, \tilde A^2, \tilde A^3)$ satisfying Requirement~\ref{ap:iid}, \ie
\[
\PP(\widetilde A^1 = j | A^1 = i) = \PP(\widetilde A^2 = j |  A^2 = i ) = \PP(\widetilde A^3 = j | A^3 = i) = T_{ij}, \forall i,j.
\]
Specially, denote by $$e_1 = \PP(\widetilde A^1 = 2 | A^1 = 1) = \PP(\widetilde A^2 = 2 | A^2 = 1) = \PP(\widetilde A^3 = 2 | A^3 = 1)$$
and
$$e_2 = \PP(\widetilde A^1 = 1 | A^1 = 2) = \PP(\widetilde A^2 = 1 | A^2 = 2) = \PP(\widetilde A^3 = 1 | A^3 = 2).$$
Note $A^1 = A^2 = A^3$.
We have:
\squishlist
    \item First order equations:
\begin{align*}
    \PP(\tilde A^1 = 1) & = \PP(A^1=1) \cdot (1-e_1) + \PP(A^1=2) \cdot  e_2  \\
    \PP(\tilde A^1 = 2) & = \PP(A^1=1) \cdot  e_1 + \PP(A^1=2) \cdot  (1-e_2) 
\end{align*}
    \item Second order equations:
\begin{align*}
    \PP(\tilde A^1 = 1, \tilde A^2 = 1) & = \PP(\tilde A^1 = 1, \tilde A^2 = 1|A^1=1) \cdot \PP(A^1=1)  + \PP(\tilde A^1 = 1, \tilde A^2 = 1|A^1=2) \cdot \PP(A^1=2) \\
    & = (1-e_1)^2 \cdot \PP(A^1=1)  + e_2^2 \cdot \PP(A^1=2).
\end{align*}
Similarly, 
\begin{align*}
    \PP(\tilde A^1 = 1, \tilde A^2 = 2) & = (1-e_1)e_1 \cdot \PP(A^1=1)  + e_2(1-e_2) \cdot \PP(A^1=2) \\
    \PP(\tilde A^1 = 2, \tilde A^2 = 1) & = (1-e_1)e_1 \cdot \PP(A^1=1)  + e_2(1-e_2) \cdot \PP(A^1=2) \\
    \PP(\tilde A^1 = 2, \tilde A^2 = 2) & = e_1^2 \cdot \PP(A^1=1)  + (1-e_2)^2 \cdot \PP(A^1=2).
\end{align*}
\item Third order equations:
\begin{align*}
    \PP(\tilde A^1 = 1, \tilde A^2 = 1, \tilde A^3 = 1) & = (1-e_1)^3 \cdot \PP(A^1=1)  + e_2^3 \cdot \PP(A^1=2) \\\
    \PP(\tilde A^1 = 1, \tilde A^2 = 1, \tilde A^3 = 2) & = (1-e_1)^2 e_1 \cdot \PP(A^1=1)  + (1-e_2)e_2^2 \cdot \PP(A^1=2) \\
    \PP(\tilde A^1 = 1, \tilde A^2 = 2, \tilde A^3 = 2) & = (1-e_1) e_1^2 \cdot \PP(A^1=1)  + (1-e_2)^2 e_2 \cdot \PP(A^1=2) \\
    \PP(\tilde A^1 = 1, \tilde A^2 = 2, \tilde A^3 = 1) & = (1-e_1)^2 e_1 \cdot \PP(A^1=1)  + (1-e_2)e_2^2 \cdot \PP(A^1=2) \\
    \PP(\tilde A^1 = 2, \tilde A^2 = 1, \tilde A^3 = 1) & = (1-e_1)^2 e_1 \cdot \PP(A^1=1)  + (1-e_2)e_2^2 \cdot \PP(A^1=2) \\
    \PP(\tilde A^1 = 2, \tilde A^2 = 1, \tilde A^3 = 2) & = (1-e_1) e_1^2 \cdot \PP(A^1=1)  + (1-e_2)^2 e_2 \cdot \PP(A^1=2) \\
    \PP(\tilde A^1 = 2, \tilde A^2 = 2, \tilde A^3 = 1) & = (1-e_1) e_1^2 \cdot \PP(A^1=1)  + (1-e_2)^2 e_2 \cdot \PP(A^1=2) \\
    \PP(\tilde A^1 = 2, \tilde A^2 = 2, \tilde A^3 = 2) & = e_1^3 \cdot \PP(A^1=1)  + (1-e_2)^3 \cdot \PP(A^1=2).
\end{align*}
\squishend

With the above equations, we can count the frequency of each pattern (LHS) as $({\hat{\vc}}^{[1]}$, ${\hat{\vc}}^{[2]}$,  ${\hat{\vc}}^{[3]})$ and solve the equations. See the key steps summarized in Algorithm~\ref{alg:key_hoc}.

\begin{algorithm}[h]
   \caption{Key Steps of HOC}
   \label{alg:key_hoc}
\begin{algorithmic}[1]
    \STATE {\bf Input:} A set of three-tuples:  $\{(\tilde a_n^1, \tilde a_n^2, \tilde a_n^3) | n\in[N]\}$ 
    \STATE $({\hat{\vc}}^{[1]}$, ${\hat{\vc}}^{[2]}$,  ${\hat{\vc}}^{[3]})\leftarrow$  \texttt{CountFreq}($\{(\tilde a_n^1, \tilde a_n^2, \tilde a_n^3) | n\in[N]\}$)  \hfill \algcom{// Count 1st, 2nd, and 3rd-order patterns}
    \STATE Find $\mT$ such that match the counts $({\hat{\vc}}^{[1]}$, ${\hat{\vc}}^{[2]}$,  ${\hat{\vc}}^{[3]})$ \hfill     \algcom{// Solve equations}
\end{algorithmic}
\end{algorithm}

\subsection{Other Estimators That Require Training}\label{appendix:estimator_training}

Many estimators~\citep{liu2015classification,scott2015rate,patrini2017making,northcutt2021confident} require extra training with target data and proxy model outputs, which introduces extra cost. Moreover, it brings a practical challenge in hyper-parameter tuning given we have no ground-truth sensitive attributes. We tried such approach but failed to get good results. 

These estimators mainly focus on training a new model to fit the noisy data distribution. The intuition is that the new model has the ability to distinguish between true attributes and wrong attributes. In other words, they believe the prediction of new model is close to the true attributes. It is useful when the noise in attributes are random. However, this intuitions is hardly true in our setting since we need to train a new model to learn the noisy attributes given by an proxy model, which are deterministic. One caveat of this approach is that the new model is likely to fit the proxy model when both the capacity of the new model and the amount of data are sufficient, leading to a trivial transition matrix estimate that is an identity matrix, i.e., $\mT = \mI$. In this case, the performance is close to \base. We reproduce \cite{northcutt2021confident} follow the setting in Table~\ref{table:celeba_diff_noise_error_full} (no additional random noise) and summarize the result in Table~\ref{tab:estimator-cl}, which verifies that the performance of this kind of approach is close to \base.

\begin{table}[!h]
    \caption{Normalized error ($\times 100$) of a learning-centric estimator.}
    \begin{center}
		\scalebox{0.79}{
    \begin{tabular}{c|cccccc}
    \toprule
 Method      & DP \globalT\   & DP \localT\ & EOd \globalT\ & EOd \localT\ & EOp \globalT\ & EOp \localT\ \\
\midrule
 \base\        & 15.33       &  /       & 4.11       &     /     & 2.82       & /         \\
 \citep{northcutt2021confident}        & 15.37       & 15.49   & 4.07      & 4.02     & 2.86       & 2.95        \\
 \bottomrule
    \end{tabular}}
    \end{center}
    \label{tab:estimator-cl}
\end{table}
\newpage

\clearpage
\newpage

\section{Full Experimental Results}\label{app:exp}

\subsection{Full Results on COMPAS}\label{app:compas_full}
We have two tables in this subsection.
\squishlist
\item Table~\ref{table:disparity_compas} shows the raw disparities measured on the COMPAS dataset.
\item Table~\ref{table:compas_full} is the full version of Table~\ref{table:compas}.
\squishend

\begin{table*}[!hb]
	\caption{Disparities in the COMPAS dataset}
	\vspace{10pt}
	\begin{center}
		\scalebox{0.89}{\footnotesize{\begin{tabular}{c|ccc ccc} 
			\hline 
			 \multirow{2}{*}{COMPAS}   & \multicolumn{3}{c}{True} & \multicolumn{3}{c}{Uncalibrated Noisy}  \\ 
			  & \DP &  \EO & \EOp & \DP &  \EO & \EOp\\
			  			\hline\hline
 tree & 0.2424 & 0.2013 & 0.2541 & 0.1362 & 0.1090 & 0.1160        \\
 forest & 0.2389 & 0.1947 & 0.2425 & 0.1346 & 0.1059 & 0.1120\\
 boosting & 0.2424 & 0.2013 & 0.2541 & 0.1362 & 0.1090 & 0.1160\\
 SVM & 0.2535 & 0.2135 & 0.2577 & 0.1252 & 0.0988 & 0.1038\\
 logit & 0.2000 & 0.1675 & 0.2278 & 0.1169 & 0.0950 & 0.1120\\
 nn & 0.2318 & 0.1913 & 0.2359 & 0.1352 & 0.1084 & 0.1073\\
 compas\_score & 0.2572 & 0.2217 & 0.2586 & 0.1511 & 0.1276 & 0.1324\\
		   \hline
		\end{tabular}}}
	\end{center}
	\label{table:disparity_compas}
\end{table*}

\begin{table*}[!h]
	\caption{Performance on the COMPAS dataset. The method with minimal normalized error is \textbf{bold}.}
	\vspace{10pt}
	\begin{center}
		\scalebox{0.83}{\footnotesize{\begin{tabular}{c|cc cc| cccc | cccc} 
			\toprule
			 \multirow{2}{*}{COMPAS}   &  \multicolumn{4}{c}{\emph{\DP\ Normalized Error (\%) $\downarrow$}} & \multicolumn{4}{c}{\emph{\EO\ Normalized Error (\%) $\downarrow$}} & \multicolumn{4}{c}{\emph{\EOp\ Normalized Error (\%) $\downarrow$}} \\
			  & \base & \soft & \globalT & \localT & \base & \soft & \globalT & \localT& \base & \soft & \globalT & \localT\\
			  			\midrule \midrule
tree & 43.82 & 61.26 & \textbf{22.29} & 39.81 & 45.86 & 63.96 & \textbf{23.09} & 42.81 & 54.36 & 70.15 & \textbf{13.27} & 49.49 \\
forest & 43.68 & 60.30 & \textbf{19.65} & 44.14 & 45.60 & 62.85 & \textbf{18.56} & 44.04 & 53.83 & 69.39 & \textbf{17.51} & 63.62 \\
boosting & 43.82 & 61.26 & \textbf{22.29} & 44.64 & 45.86 & 63.96 & \textbf{23.25} & 49.08 & 54.36 & 70.15 & \textbf{13.11} & 54.67 \\
SVM & 50.61 & 66.50 & \textbf{30.95} & 42.00 & 53.72 & 69.69 & \textbf{32.46} & 47.39 & 59.70 & 71.12 & \textbf{29.29} & 51.31 \\
logit & 41.54 & 60.78 & \textbf{16.98} & 35.69 & 43.26 & 63.15 & \textbf{21.42} & 31.91 & 50.86 & 65.04 & \textbf{14.90} & 26.27 \\
nn & 41.69 & 60.55 & \textbf{19.48} & 34.22 & 43.34 & 62.99 & \textbf{19.30} & 43.24 & 54.50 & 68.50 & \textbf{14.20} & 59.95 \\
compas\_score & 41.28 & 58.34 & \textbf{11.24} & 14.66 & 42.43 & 59.79 & \textbf{11.80} & 18.65 & 48.78 & 62.24 & \textbf{5.78} & 23.80 \\
		  \midrule \midrule
 &  \multicolumn{4}{c}{\emph{\DP\ Raw Disparity  $\downarrow$}} & \multicolumn{4}{c}{\emph{\EO\ Raw Disparity  $\downarrow$}} & \multicolumn{4}{c}{\emph{\EOp\ Raw Disparity  $\downarrow$}} \\
			  			\midrule
tree & 0.1362 & 0.0939 & 0.1884 & 0.1459 & 0.1090 & 0.0726 & 0.1548 & 0.1151 & 0.1160 & 0.0759 & 0.2204 & 0.1283 \\
forest & 0.1345 & 0.0948 & 0.1919 & 0.1334 & 0.1059 & 0.0723 & 0.1586 & 0.1090 & 0.1120 & 0.0743 & 0.2001 & 0.0882 \\
boosting & 0.1362 & 0.0939 & 0.1884 & 0.1342 & 0.1090 & 0.0726 & 0.1545 & 0.1025 & 0.1160 & 0.0759 & 0.2208 & 0.1152 \\
SVM & 0.1252 & 0.0849 & 0.1750 & 0.1470 & 0.0988 & 0.0647 & 0.1442 & 0.1123 & 0.1038 & 0.0744 & 0.1822 & 0.1255 \\
logit & 0.1169 & 0.0784 & 0.1660 & 0.1286 & 0.0950 & 0.0617 & 0.1316 & 0.1140 & 0.1120 & 0.0797 & 0.1939 & 0.1680 \\
nn & 0.1352 & 0.0915 & 0.1867 & 0.1525 & 0.1084 & 0.0708 & 0.1544 & 0.1086 & 0.1073 & 0.0743 & 0.2024 & 0.0945 \\
compas\_score & 0.1510 & 0.1072 & 0.2283 & 0.2195 & 0.1276 & 0.0891 & 0.1955 & 0.1803 & 0.1324 & 0.0976 & 0.2436 & 0.1970 \\
		  \midrule \midrule
 &  \multicolumn{4}{c}{\emph{\DP\ Raw Error  $\downarrow$}} & \multicolumn{4}{c}{\emph{\EO\ Raw Error  $\downarrow$}} & \multicolumn{4}{c}{\emph{\EOp\ Raw Error  $\downarrow$}} \\
			  			\midrule
tree & 0.1062 & 0.1485 & 0.0540 & 0.0965 & 0.0923 & 0.1288 & 0.0465 & 0.0862 & 0.1381 & 0.1782 & 0.0337 & 0.1257 \\
forest & 0.1043 & 0.1440 & 0.0469 & 0.1054 & 0.0888 & 0.1224 & 0.0361 & 0.0858 & 0.1306 & 0.1683 & 0.0425 & 0.1543 \\
boosting & 0.1062 & 0.1485 & 0.0540 & 0.1082 & 0.0923 & 0.1288 & 0.0468 & 0.0988 & 0.1381 & 0.1782 & 0.0333 & 0.1389 \\
SVM & 0.1283 & 0.1685 & 0.0785 & 0.1064 & 0.1147 & 0.1488 & 0.0693 & 0.1012 & 0.1538 & 0.1833 & 0.0755 & 0.1322 \\
logit & 0.0831 & 0.1215 & 0.0340 & 0.0714 & 0.0724 & 0.1057 & 0.0359 & 0.0534 & 0.1159 & 0.1482 & 0.0339 & 0.0598 \\
nn & 0.0966 & 0.1404 & 0.0452 & 0.0793 & 0.0829 & 0.1205 & 0.0369 & 0.0827 & 0.1286 & 0.1616 & 0.0335 & 0.1414 \\
compas\_score & 0.1062 & 0.1500 & 0.0289 & 0.0377 & 0.0941 & 0.1325 & 0.0261 & 0.0413 & 0.1261 & 0.1609 & 0.0150 & 0.0615 \\
		  \midrule \midrule
 &  \multicolumn{4}{c}{\emph{\DP\ Improvement (\%) $\uparrow$}} & \multicolumn{4}{c}{\emph{\EO\ Improvement (\%) $\uparrow$}} & \multicolumn{4}{c}{\emph{\EOp\ Improvement (\%) $\uparrow$}} \\
			  			\midrule
tree & 0.00 & -39.79 & 49.15 & 9.15 & 0.00 & -39.48 & 49.65 & 6.64 & 0.00 & -29.05 & 75.60 & 8.96 \\
forest & 0.00 & -38.05 & 55.01 & -1.06 & 0.00 & -37.83 & 59.30 & 3.42 & 0.00 & -28.89 & 67.47 & -18.18 \\
boosting & 0.00 & -39.79 & 49.15 & -1.87 & 0.00 & -39.48 & 49.30 & -7.04 & 0.00 & -29.05 & 75.89 & -0.57 \\
SVM & 0.00 & -31.40 & 38.83 & 17.02 & 0.00 & -29.72 & 39.57 & 11.78 & 0.00 & -19.12 & 50.93 & 14.05 \\
logit & 0.00 & -46.30 & 59.12 & 14.08 & 0.00 & -45.98 & 50.47 & 26.24 & 0.00 & -27.87 & 70.70 & 48.35 \\
nn & 0.00 & -45.23 & 53.27 & 17.93 & 0.00 & -45.34 & 55.47 & 0.23 & 0.00 & -25.69 & 73.94 & -10.01 \\
compas\_score & 0.00 & -41.33 & 72.77 & 64.48 & 0.00 & -40.92 & 72.20 & 56.04 & 0.00 & -27.59 & 88.15 & 51.21 \\
		   \bottomrule
		\end{tabular}}}
	\end{center}
	\label{table:compas_full}
\end{table*}

\newpage

\subsection{Experiments on COMPAS With Three-Class Sensitive Attributes}\label{appendix:compas_3}

We experiment with three categories of sensitive attributes: black, white, and others, and show the result in Table~\ref{table:compas_3}. 
Table~\ref{table:compas_3} shows our proposed algorithm with global estimates is consistently and significantly better than the baselines, which is also consistent with the results from Table~\ref{table:compas}.

\begin{table*}[!hb]
	\caption{Normalized estimation error on COMPAS. Each row is a different target model $f$.}
	\vspace{10pt}
	\begin{center}
		\scalebox{0.89}{\footnotesize{\begin{tabular}{c|cc cc| cccc | cccc} 
			\toprule
			 COMPAS  &  \multicolumn{4}{c}{\emph{\DP\ Normalized Error (\%) $\downarrow$}} & \multicolumn{4}{c}{\emph{\EO\ Normalized Error (\%) $\downarrow$}} & \multicolumn{4}{c}{\emph{\EOp\ Normalized Error (\%) $\downarrow$}}  \\
			 \textit{True disparity:} $\sim 0.2$ & \base & \soft & \globalT & \localT & \base & \soft & \globalT & \localT\\
			  			\midrule \midrule
tree & 24.87 & 59.98 & \textbf{13.84} & 25.16 & 30.15 & 63.13 & \textbf{13.11} & 27.84 & 42.42 & 68.50 & \textbf{~~4.46} & 43.54 \\
forest & 23.94 & 58.67 & \textbf{10.00} & 26.64 & 29.19 & 61.66 & \textbf{11.61} & 33.85 & 41.53 & 67.50 & \textbf{~~1.77} & 45.03\\
boosting & 24.87 & 59.98 & \textbf{13.84} & 25.44 & 30.15 & 63.13 & \textbf{15.74} & 33.20 & 42.42 & 68.50 & \textbf{~~6.19} & 47.85\\
SVM & 40.37 & 67.02 & \textbf{25.96} & 34.73 & 49.57 & 71.56 & \textbf{29.33} & 42.91 & 56.55 & 73.66 & \textbf{17.69} & 37.74 \\
logit & 16.71 & 58.46 & \textbf{~~7.39} & 22.17 & 17.23 & 60.64 & \textbf{~~7.02} & 25.38 & 22.24 & 59.77 & \textbf{13.48} & 26.13 \\
nn & 18.60 & 58.05 & \textbf{~~5.38} & 16.58 & 22.91 & 61.42 & \textbf{~~5.90} & 22.63 & 33.55 & 65.23 & \textbf{~~0.94} & 45.84\\
compas\_score & 29.00 & 59.17 & \textbf{10.02} & 31.32 & 33.43 & 62.05 & \textbf{12.15} & 36.03 & 39.93 & 65.31 & \textbf{~~4.38} & 44.82\\
		   \bottomrule
		\end{tabular}}}
	\end{center}
	\label{table:compas_3}
\end{table*}

\subsection{Full Results on CelebA}\label{app:celeba_full}
We have two tables in this subsection.
\squishlist
\item Table~\ref{table:celeba_diff_noise_error_full} is the full version of Table~\ref{table:celebA_noise_facenet512}.
\item Table~\ref{table:celeba_diff_noise_improve_full} is similar to Table~\ref{table:celeba_diff_noise_error_full}, but the error metric is changed to \textsf{Improvement} defined in Section~\ref{sec:exp_setting}.
\squishend

\begin{table*}[!ht]
	\caption{Normalized Error on CelebA with different noise rates}
	\vspace{10pt}
	\begin{center}
		\scalebox{0.82}{\footnotesize{\begin{tabular}{c|cc cc| cccc | cccc} 
			\toprule
			 \multirow{2}{*}{CelebA}   &  \multicolumn{4}{c}{\emph{\DP\ Normalized Error (\%) $\downarrow$}} & \multicolumn{4}{c}{\emph{\EO\ Normalized Error (\%) $\downarrow$}} & \multicolumn{4}{c}{\emph{\EOp\ Normalized Error (\%) $\downarrow$}} \\
			  & \base & \soft & \globalT & \localT & \base & \soft & \globalT & \localT& \base & \soft & \globalT & \localT\\
			  			\midrule \midrule

Facenet [0.0, 0.0] & 15.33 & 12.54 & 22.17 & 10.89 & 4.11 & 6.46 & 7.54 & 0.26 & 2.82 & 0.34 & 12.22 & 2.93 \\
    Facenet [0.2, 0.0] & 7.39 & 11.65 & 20.75 & 10.82 & 25.05 & 26.99 & 9.87 & 6.63 & 24.69 & 27.27 & 11.55 & 2.77 \\
    Facenet [0.2, 0.2] & 30.24 & 31.57 & 24.27 & 8.45 & 44.71 & 46.36 & 15.10 & 3.99 & 37.67 & 38.77 & 21.79 & 16.73 \\
    Facenet [0.4, 0.2] & 51.37 & 54.56 & 20.12 & 20.66 & 62.94 & 65.10 & 3.45 & 3.67 & 56.53 & 58.73 & 15.75 & 2.70 \\
    Facenet [0.4, 0.4] & 77.82 & 78.39 & 8.76 & 21.94 & 79.36 & 80.10 & 51.32 & 148.05 & 78.39 & 79.62 & 71.38 & 146.20 \\ \hline
Facenet512 [0.0, 0.0] & 15.33 & 12.54 & 21.70 & 7.26 & 4.11 & 6.46 & 4.85 & 0.52 & 2.82 & 0.34 & 11.80 & 3.24 \\
    Facenet512 [0.2, 0.0] & 7.37 & 11.65 & 20.58 & 5.05 & 25.06 & 26.99 & 6.43 & 0.10 & 24.69 & 27.27 & 11.11 & 1.07 \\
    Facenet512 [0.2, 0.2] & 30.21 & 31.57 & 24.25 & 13.10 & 44.73 & 46.36 & 11.26 & 9.04 & 37.67 & 38.77 & 20.94 & 27.98 \\
    Facenet512 [0.4, 0.2] & 51.32 & 54.56 & 19.42 & 10.47 & 62.90 & 65.10 & 11.09 & 19.15 & 56.51 & 58.73 & 23.86 & 23.55 \\
    Facenet512 [0.4, 0.4] & 77.76 & 78.39 & 9.41 & 19.80 & 79.31 & 80.10 & 24.49 & 8.02 & 78.35 & 79.62 & 10.61 & 5.71 \\ \hline
OpenFace [0.0, 0.0] & 15.33 & 12.54 & 10.31 & 9.39 & 4.11 & 6.46 & 10.43 & 5.03 & 2.82 & 0.34 & 0.56 & 0.93 \\
    OpenFace [0.2, 0.0] & 7.39 & 11.65 & 8.93 & 6.60 & 25.05 & 26.99 & 9.86 & 13.01 & 24.69 & 27.27 & 1.08 & 10.96 \\
    OpenFace [0.2, 0.2] & 30.24 & 31.57 & 13.32 & 21.46 & 44.74 & 46.36 & 7.56 & 15.88 & 37.69 & 38.77 & 5.90 & 7.40 \\
    OpenFace [0.4, 0.2] & 51.39 & 54.56 & 10.66 & 25.16 & 62.96 & 65.10 & 6.47 & 24.94 & 56.55 & 58.73 & 6.11 & 47.12 \\
    OpenFace [0.4, 0.4] & 77.84 & 78.39 & 1.60 & 117.27 & 79.38 & 80.10 & 34.00 & 19.47 & 78.41 & 79.62 & 37.42 & 31.99 \\ \hline
ArcFace [0.0, 0.0] & 15.33 & 12.54 & 19.59 & 9.69 & 4.11 & 6.46 & 5.72 & 0.23 & 2.82 & 0.34 & 11.16 & 3.85 \\
    ArcFace [0.2, 0.0] & 7.39 & 11.65 & 17.74 & 7.74 & 25.05 & 26.99 & 6.18 & 1.82 & 24.69 & 27.27 & 8.81 & 3.37 \\
    ArcFace [0.2, 0.2] & 30.19 & 31.57 & 21.77 & 8.97 & 44.77 & 46.36 & 12.12 & 18.91 & 37.69 & 38.77 & 21.19 & 17.99 \\
    ArcFace [0.4, 0.2] & 51.32 & 54.56 & 17.33 & 44.52 & 62.91 & 65.10 & 14.66 & 29.74 & 56.53 & 58.73 & 24.39 & 4.92 \\
    ArcFace [0.4, 0.4] & 77.79 & 78.39 & 8.38 & 84.37 & 79.34 & 80.10 & 8.31 & 165.03 & 78.39 & 79.62 & 16.98 & 62.34 \\ \hline
Dlib [0.0, 0.0] & 15.33 & 12.54 & 15.09 & 5.30 & 4.11 & 6.46 & 4.87 & 4.25 & 2.82 & 0.34 & 9.74 & 2.32 \\
    Dlib [0.2, 0.0] & 7.35 & 11.65 & 14.39 & 1.06 & 25.07 & 26.99 & 3.78 & 2.63 & 24.69 & 27.27 & 7.09 & 2.36 \\
    Dlib [0.2, 0.2] & 30.23 & 31.57 & 16.78 & 1.95 & 44.77 & 46.36 & 9.50 & 11.28 & 37.72 & 38.77 & 15.88 & 22.43 \\
    Dlib [0.4, 0.2] & 51.40 & 54.56 & 12.83 & 17.69 & 62.96 & 65.10 & 10.34 & 11.47 & 56.57 & 58.73 & 18.90 & 11.17 \\
    Dlib [0.4, 0.4] & 77.84 & 78.39 & 0.46 & 96.58 & 79.38 & 80.10 & 7.99 & 86.36 & 78.41 & 79.62 & 8.45 & 14.78 \\ \hline
SFace [0.0, 0.0] & 15.33 & 12.54 & 17.00 & 4.77 & 4.11 & 6.46 & 4.04 & 3.91 & 2.82 & 0.34 & 9.36 & 3.28 \\
    SFace [0.2, 0.0] & 7.41 & 11.65 & 15.18 & 1.94 & 25.04 & 26.99 & 3.31 & 8.82 & 24.69 & 27.27 & 7.24 & 13.05 \\
    SFace [0.2, 0.2] & 30.22 & 31.57 & 18.16 & 20.95 & 44.72 & 46.36 & 4.58 & 20.93 & 37.67 & 38.77 & 11.55 & 34.72 \\
    SFace [0.4, 0.2] & 51.35 & 54.56 & 14.72 & 48.96 & 62.92 & 65.10 & 2.95 & 68.93 & 56.51 & 58.73 & 15.22 & 68.85 \\
    SFace [0.4, 0.4] & 77.78 & 78.39 & 3.37 & 31.25 & 79.33 & 80.10 & 21.56 & 178.21 & 78.37 & 79.62 & 20.03 & 86.59 \\ \hline
		   \bottomrule
		\end{tabular}}}
	\end{center}
	\label{table:celeba_diff_noise_error_full}
\end{table*}

\begin{table*}[!ht]
	\caption{Improvement on CelebA with different noise rates}
	\vspace{10pt}
	\begin{center}
		\scalebox{0.82}{\footnotesize{\begin{tabular}{c|cc cc| cccc | cccc} 
			\toprule
			 \multirow{2}{*}{CelebA}   &  \multicolumn{4}{c}{\emph{\DP\ Improvement (\%) $\uparrow$}} & \multicolumn{4}{c}{\emph{\EO\ Improvement (\%) $\uparrow$}} & \multicolumn{4}{c}{\emph{\EOp\ Improvement (\%) $\uparrow$}} \\
			  & \base & \soft & \globalT & \localT & \base & \soft & \globalT & \localT& \base & \soft & \globalT & \localT\\
			  			\midrule \midrule
Facenet [0.0, 0.0] & 0.00 & 18.22 & \color{black}{-44.58} & 28.99 & 0.00 & \color{black}{-57.38} & \color{black}{-83.62} & 93.64 & 0.00 & 88.05 & \color{black}{-333.85} & \color{black}{-3.97} \\
    Facenet [0.2, 0.0] & 0.00 & \color{black}{-57.70} & \color{black}{-180.88} & \color{black}{-46.45} & 0.00 & \color{black}{-7.75} & 60.60 & 73.52 & 0.00 & \color{black}{-10.44} & 53.24 & 88.80 \\
    Facenet [0.2, 0.2] & 0.00 & \color{black}{-4.39} & 19.75 & 72.05 & 0.00 & \color{black}{-3.69} & 66.22 & 91.07 & 0.00 & \color{black}{-2.92} & 42.17 & 55.58 \\
    Facenet [0.4, 0.2] & 0.00 & \color{black}{-6.20} & 60.83 & 59.79 & 0.00 & \color{black}{-3.44} & 94.51 & 94.17 & 0.00 & \color{black}{-3.90} & 72.13 & 95.23 \\
    Facenet [0.4, 0.4] & 0.00 & \color{black}{-0.73} & 88.74 & 71.81 & 0.00 & \color{black}{-0.94} & 35.33 & \color{black}{-86.56} & 0.00 & \color{black}{-1.57} & 8.94 & \color{black}{-86.50} \\ \hline
Facenet512 [0.0, 0.0] & 0.00 & 18.22 & \color{black}{-41.50} & 52.65 & 0.00 & \color{black}{-57.38} & \color{black}{-18.15} & 87.29 & 0.00 & 88.05 & \color{black}{-319.18} & \color{black}{-15.09} \\
    Facenet512 [0.2, 0.0] & 0.00 & \color{black}{-58.10} & \color{black}{-179.28} & 31.43 & 0.00 & \color{black}{-7.70} & 74.32 & 99.58 & 0.00 & \color{black}{-10.44} & 54.98 & 95.68 \\
    Facenet512 [0.2, 0.2] & 0.00 & \color{black}{-4.51} & 19.72 & 56.64 & 0.00 & \color{black}{-3.64} & 74.81 & 79.78 & 0.00 & \color{black}{-2.92} & 44.40 & 25.73 \\
    Facenet512 [0.4, 0.2] & 0.00 & \color{black}{-6.32} & 62.17 & 79.60 & 0.00 & \color{black}{-3.50} & 82.37 & 69.55 & 0.00 & \color{black}{-3.94} & 57.78 & 58.33 \\
    Facenet512 [0.4, 0.4] & 0.00 & \color{black}{-0.81} & 87.90 & 74.54 & 0.00 & \color{black}{-1.00} & 69.12 & 89.89 & 0.00 & \color{black}{-1.63} & 86.45 & 92.71 \\ \hline
OpenFace [0.0, 0.0] & 0.00 & 18.22 & 32.76 & 38.75 & 0.00 & \color{black}{-57.38} & \color{black}{-154.12} & \color{black}{-22.45} & 0.00 & 88.05 & 80.03 & 67.15 \\
    OpenFace [0.2, 0.0] & 0.00 & \color{black}{-57.70} & \color{black}{-20.83} & 10.69 & 0.00 & \color{black}{-7.75} & 60.65 & 48.05 & 0.00 & \color{black}{-10.44} & 95.64 & 55.62 \\
    OpenFace [0.2, 0.2] & 0.00 & \color{black}{-4.38} & 55.97 & 29.06 & 0.00 & \color{black}{-3.62} & 83.11 & 64.51 & 0.00 & \color{black}{-2.86} & 84.35 & 80.38 \\
    OpenFace [0.4, 0.2] & 0.00 & \color{black}{-6.16} & 79.25 & 51.05 & 0.00 & \color{black}{-3.41} & 89.72 & 60.39 & 0.00 & \color{black}{-3.86} & 89.19 & 16.67 \\
    OpenFace [0.4, 0.4] & 0.00 & \color{black}{-0.71} & 97.94 & \color{black}{-50.65} & 0.00 & \color{black}{-0.92} & 57.17 & 75.47 & 0.00 & \color{black}{-1.54} & 52.28 & 59.20 \\ \hline
ArcFace [0.0, 0.0] & 0.00 & 18.22 & \color{black}{-27.78} & 36.78 & 0.00 & \color{black}{-57.38} & \color{black}{-39.45} & 94.31 & 0.00 & 88.05 & \color{black}{-296.25} & \color{black}{-36.65} \\
    ArcFace [0.2, 0.0] & 0.00 & \color{black}{-57.70} & \color{black}{-140.07} & \color{black}{-4.72} & 0.00 & \color{black}{-7.75} & 75.31 & 92.72 & 0.00 & \color{black}{-10.44} & 64.32 & 86.37 \\
    ArcFace [0.2, 0.2] & 0.00 & \color{black}{-4.56} & 27.91 & 70.28 & 0.00 & \color{black}{-3.55} & 72.94 & 57.76 & 0.00 & \color{black}{-2.86} & 43.79 & 52.27 \\
    ArcFace [0.4, 0.2] & 0.00 & \color{black}{-6.31} & 66.22 & 13.25 & 0.00 & \color{black}{-3.49} & 76.69 & 52.72 & 0.00 & \color{black}{-3.90} & 56.85 & 91.29 \\
    ArcFace [0.4, 0.4] & 0.00 & \color{black}{-0.78} & 89.23 & \color{black}{-8.47} & 0.00 & \color{black}{-0.97} & 89.53 & \color{black}{-108.01} & 0.00 & \color{black}{-1.57} & 78.34 & 20.47 \\ \hline
Dlib [0.0, 0.0] & 0.00 & 18.22 & 1.56 & 65.46 & 0.00 & \color{black}{-57.38} & \color{black}{-18.55} & \color{black}{-3.43} & 0.00 & 88.05 & \color{black}{-245.95} & 17.61 \\
    Dlib [0.2, 0.0] & 0.00 & \color{black}{-58.50} & \color{black}{-95.79} & 85.62 & 0.00 & \color{black}{-7.66} & 84.90 & 89.53 & 0.00 & \color{black}{-10.44} & 71.30 & 90.42 \\
    Dlib [0.2, 0.2] & 0.00 & \color{black}{-4.43} & 44.49 & 93.54 & 0.00 & \color{black}{-3.53} & 78.78 & 74.80 & 0.00 & \color{black}{-2.80} & 57.89 & 40.54 \\
    Dlib [0.4, 0.2] & 0.00 & \color{black}{-6.15} & 75.03 & 65.59 & 0.00 & \color{black}{-3.39} & 83.58 & 81.78 & 0.00 & \color{black}{-3.82} & 66.59 & 80.25 \\
    Dlib [0.4, 0.4] & 0.00 & \color{black}{-0.71} & 99.41 & \color{black}{-24.07} & 0.00 & \color{black}{-0.92} & 89.94 & \color{black}{-8.80} & 0.00 & \color{black}{-1.54} & 89.22 & 81.15 \\ \hline
SFace [0.0, 0.0] & 0.00 & 18.22 & \color{black}{-10.87} & 68.87 & 0.00 & \color{black}{-57.38} & 1.61 & 4.85 & 0.00 & 88.05 & \color{black}{-232.48} & \color{black}{-16.46} \\
    SFace [0.2, 0.0] & 0.00 & \color{black}{-57.31} & \color{black}{-104.91} & 73.84 & 0.00 & \color{black}{-7.79} & 86.78 & 64.75 & 0.00 & \color{black}{-10.44} & 70.66 & 47.12 \\
    SFace [0.2, 0.2] & 0.00 & \color{black}{-4.45} & 39.93 & 30.68 & 0.00 & \color{black}{-3.67} & 89.76 & 53.18 & 0.00 & \color{black}{-2.92} & 69.34 & 7.82 \\
    SFace [0.4, 0.2] & 0.00 & \color{black}{-6.24} & 71.34 & 4.66 & 0.00 & \color{black}{-3.47} & 95.32 & \color{black}{-9.55} & 0.00 & \color{black}{-3.94} & 73.06 & \color{black}{-21.85} \\
    SFace [0.4, 0.4] & 0.00 & \color{black}{-0.78} & 95.67 & 59.82 & 0.00 & \color{black}{-0.98} & 72.82 & \color{black}{-124.64} & 0.00 & \color{black}{-1.60} & 74.44 & \color{black}{-10.49} \\
		   \bottomrule
		\end{tabular}}}
	\end{center}
	\label{table:celeba_diff_noise_improve_full}
\end{table*}

\subsection{Discussions on \globalT}\label{appendix:discuss_global}

\globalT\ is a heuristic to better estimate $\bm T_k$ when $\mT_k$ cannot be estimated stably. According to Theorem~\ref{thm:error_correction}, when $\mT_k$s are accurately estimated, we should always rely on the local estimates as Line~\ref{alg_line:localT} of Algorithm~\ref{alg:alg_hoc} to achieve a zero calibration error. However, in practice, each time when we estimate a local $\widehat{\mT}_{k}$, the estimator would introduce certain error on the $\widehat{\mT}_{k}$ and the matrix inversion in Theorem~\ref{lem:relation_true_noisy} might amplify the estimation error on $\widehat{\mT}_{k}$ each time, leading to a large overall error on the metric. One \textit{heuristic} is to use a single global transition matrix $\widehat{\mT}$ estimated once on the full dataset $\widetilde D$ as Line~8 of Algorithm~\ref{alg:alg_hoc} to approximate ${\mT}_{k}$. Intuitively, $\widehat{\mT}$ can be viewed as the weighted average of all $\widehat{\mT}_{k}$'s to stabilize estimation error (variance reduction) on $\widehat{\mT}_{k}$. Admittedly, the average will introduce bias since the equation in Theorem~\ref{lem:relation_true_noisy} would not hold when replacing $\mT_{k}$ with $\mT$. The justification is that the error introduced by violating the equality might be smaller than the error introduced by using severely inaccurately estimates of $\mT_{k}$'s. Therefore, we offer two options for estimating $\bm T_k$ in practice: locals estimates $\bm T_k \approx \widehat{\bm T}_k$ and global estimates $\bm T_k \approx \widehat{\bm T}$. Although it is hard to guarantee which option must be better in reality, we report the experimental results using both options and provide insights for choosing between both estimates in Sec.~\ref{subsec:exp_results}.

\subsection{Disparity Mitigation With Our Calibration Algorithm}\label{appendix:mitigation}

We apply our calibration algorithm to mitigate disparity during training. Specifically, the local method is applied on the CelebA dataset. The preprocess of the dataset and generation of noisy sensitive attributes are the same as the experiments in Table 2. The backbone network is ViT-B\_8 \citep{dosovitskiy2020image}. The aim is to improve the classification accuracy while ensuring \DP, where $\widehat \Delta(\widetilde D,f) = 0$ is the constraint during training. 
Specifically, the optimization problem is
\begin{align*}
    \min_{f} &\quad \sum_{n=1}^N \ell(f(x_n),y_n) \\
    \it{s.t.} & \quad \widehat \Delta(\widetilde D,f) = 0,
\end{align*}
where $\ell$ is the cross-entropy loss.
Recall $\widehat \Delta(\widetilde D,f)$ is obtained from our Algorithm 1 (Line 8), and $\widetilde D:=\{(x_n,y_n,\tilde a_n) | n\in[N]\}$. Noting the constraint is not differentiable since it depends on the sample counts, i.e., $$\widetilde H[\tilde a,k] = \PP(f(X)=k|\widetilde A=\tilde a) \approx \frac{1}{N} \sum_{n=1}^N\BR(f(x_n=k|\tilde a_n=\tilde a)).$$
To make it differentiable, we use a relaxed measure \citep{madras2018learning,wang2022understanding} as follows:
$$\widetilde H[\tilde a,k] = \PP(f(X)=k|\widetilde A=\tilde a) \approx \frac{1}{N_{\tilde a}} \sum_{n=1, \tilde a_n = \tilde a}^N \vf_{x_n}[k],$$
where $\vf_{x_n}[k]$ is the model's prediction probability on class $k$, and $N_{\tilde a}$ is the number of samples that have noisy attribute $\tilde a$.
The standard method of multipliers is employed to train with constraints \citep{boyd2011distributed}.
We train the model for 20 epochs with a stepsize of 256. Table~\ref{table:mitigation} shows the accuracy and DP disparity on the test data averaged with results from the last 5 epochs of training. From the table, we conclude that, with any selected pre-trained model, the mitigation based on our calibration results significantly outperforms the direct mitigation with noisy attributes in terms of both accuracy improvement and disparity mitigation.

\end{document}